\pdfoutput=1
\documentclass[final]{cvpr}

\usepackage{times}
\usepackage{epsfig}
\usepackage{graphicx}
\usepackage{amsmath,amssymb, amsthm, amsfonts, mathtools, multirow}
\usepackage{array,booktabs}
\usepackage{float}
\usepackage{nopageno}

\newcommand{\bm}[2]{\rule{0pt}{4ex}$\begin{bmatrix} {#1}\\{#2} \end{bmatrix}$}
\newcommand{\bmm}[3]{\rule{0pt}{3.5ex}$\begin{bmatrix} {#1}\\{#2}\\{#3} \end{bmatrix}$}

\newcolumntype{M}[1]{>{\centering\arraybackslash}m{#1}}
\newcommand{\vc}[3]{\overset{#3}{\underset{#2}{#1}}}

\usepackage[pagebackref=true,breaklinks=true,colorlinks,bookmarks=false]{hyperref}

\def\cvprPaperID{13} 
\def\confYear{CVPR 2021}

\begin{document}

\title{A Theoretical‐Empirical Approach to Estimating Sample Complexity of DNNs}

\author{Devansh Bisla\thanks{Equal Contribution}, \,\, Apoorva Nandini Saridena\footnotemark[1], \,\,Anna Choromanska \\
Department of Electrical and Computer Engineering, \\ 
Tandon School of Engineering, New York University \\
{\tt\small \{db3484, ans609, ac5455\}@nyu.edu}
}

\maketitle

\begin{abstract}
\vspace{-0.15in}
This paper focuses on understanding how the generalization error scales with the amount of the training data for deep neural networks (DNNs). Existing techniques in statistical learning theory require a computation of capacity measures, such as VC dimension, to provably bound this error. It is however unclear how to extend these measures to DNNs and therefore the existing analyses are applicable to simple neural networks, which are not used in practice, e.g., linear or shallow (at most two-layer) ones or otherwise multi-layer perceptrons. Moreover many theoretical error bounds are not empirically verifiable. In this paper we derive estimates of the generalization error that hold for deep networks and do not rely on unattainable capacity measures. The enabling technique in our approach hinges on two major assumptions: i) the network achieves zero training error, ii) the probability of making an error on a test point is proportional to the distance between this point and its nearest training point in the feature space and at certain maximal distance (that we call radius) it saturates. Based on these assumptions we estimate the generalization error of DNNs. The obtained estimate scales as $\mathcal{O}\left(\frac{1}{\delta N^{1/d}}\right)$, where $N$ is the size of the training data, and is parameterized by two quantities, the effective dimensionality of the data as perceived by the network ($d$) and the aforementioned radius ($\delta$), both of which we find empirically. We show that our estimates match with the experimentally-obtained behavior of the error on multiple learning tasks using benchmark data-sets and realistic models. Estimating training data requirements is essential for deployment of safety critical applications such as autonomous driving, medical diagnostics etc. Furthermore, collecting and annotating training data requires a huge amount of financial, computational and human resources. Our empirical estimates will help to efficiently allocate resources.
\vspace{-0.25in}
\end{abstract}

\section{Introduction}
Deep learning (DL) establishes state-of-the-art performances in a number of learning tasks such as image recognition~\cite{russakovsky2015imagenet,NIPS2012_4824}, speech recognition~\cite{alex2018speech,chiu2018state}, and natural language processing~\cite{Bahdanau2014nlp,DBLP:journals/corr/abs-1810-04805}. In recent years DL approaches were also shown to outperform humans in classic games such as Go~\cite{Silver_2016}. The performance of DL models depends on three factors: the model's architecture, data set, and training infrastructure, which together constitute the artificial intelligence (AI) trinity framework~\cite{Trinity}). It has been observed empirically~\cite{chen2019data} that increasing size of the training data is an extremely effective way for improving the performance of DL models, more so than modifying the network's architecture or training infrastructure. Generalization ability of a network is defined as difference between the training and test performance of the network. The relationship between the generalization ability of DL model and the size of the training data set constitutes a fundamental characteristic of the AI trinity framework, yet it is poorly described in the literature. Addressing this problem is crucial for safety critical applications where it is desired to understand the sample complexity of the model, or in other words, estimate the amount of data needed to achieve a certain acceptable level of performance. This paper aims at describing this relationship, in the context of a supervised learning setting, through a set of mixed mathematical-empirical tools.

Classic approaches in statistical learning theory for analyzing sample complexity rely on the measures of capacity of a classification model, such Vapnik-Chervonenkis (VC) dimension~\cite{vapnik2013nature} or Rademacher complexity~\cite{bartlett2002rademacher}, which are potentially prohibitive to extend to practical DL models and may lead to loose estimates (for example, in the case where these measures approach infinity). The aim of this work is to develop a framework to model the sample complexity of DNNs that is free from such measures and straightforward to use by practitioners. We propose to model the probability of making an error by a DNN on a test sample as a function of the distance in the feature space between that sample (we denote the feature vector of the test sample as $\hat{x}$) and its nearest training sample (we denote the feature vector of this training sample as $x(\hat{x})$). This probability takes the following form:
\vspace{-10pt}
\begin{equation}
    \label{eq:phi_of_x}
    \Phi(\hat{x}) \coloneqq \min\Big(1, \frac{||\hat{x} - x(\hat{x})||_2}{\delta}\Big),
\end{equation}
where $\delta$ is a positive constant that we call the radius\footnote{Our empirical results were not sensitive to the choice of distance measure. We obtained similar results for squared distance.}. Thus, when test and training examples are further than $\delta$ in the feature space, the test example will be misclassified according to our model. The choice of the form of $\Phi(\hat{x})$ is also intuitive since as we increase the amount of training data, it becomes more reasonable to expect a test point within a certain distance from the training data point. Our approach is motivated by the fact that fundamentally DNNs can be interpreted as methods of non-linear dimensionality reduction which cluster together input data with similar features while push further apart dis-similar data points~\cite{amjad2019learning,Yang_2016_CVPR}. Therefore, a test point far from the closest training point in the feature space is more likely to be incorrectly classified. In addition to assuming the aforementioned error mechanism for DNNs, we also assume that DNNs under consideration can learn perfectly and achieve zero error on the training data set. This assumption is in practice non-restrictive as it was observed empirically that DNNs of sufficient size can fit accurately labelled training data.\footnote{We assume that labeling of the training data is consistent and does not contain any mistakes.} Finally, in order to estimate the generalization error in the function of the training data size ($N$), we define the notion of model-dependent effective dimensionality of the data $d$. It is a minimum dimensionality to which one can compress the feature vector without affecting the performance of the model. We find that this dimensionality is very low in practice. Under the above framework we estimate that the generalization error of a DNN follows a power law and scales inversely to $\delta N^{1/d}$. To the best of our knowledge, our work is distinct from  existing work in the literature on the sample complexity of DNNs in a number of ways: i) it provides generalization error estimates, rather than mathematically-rigid error bounds ii) our result is free from complex capacity measures and relies on quantities ($d$ and $\delta$) which can be easily obtained empirically, iii) we perform exhaustive experimental evaluation of our theoretical result, and iv) our estimates can be easily adopted by  practitioners to asses the amount of training data needed to meet the performance requirements of their applications.

The paper is organized as follows; Section~\ref{sec:related_work} reviews the related work,  Section~\ref{sec:theoretical_framework} provides the mathematical derivations, Section~\ref{sec:experiments} reports the experimental results, and finally Section~\ref{sec:conclusion} concludes the paper. The Supplement contains additional mathematical derivations and empirical evidence.

\section{Background and Related Works}
\label{sec:related_work}
Statistical learning theory typically bounds the generalization error~\cite{bousquet2003introduction,doi:10.1002/wics.179,jiang2020fantastic} using concentration inequalities, e.g., Hoeffding's inequality~\cite{hoeffding1994probability}. The error bounds depend on the measure of the complexity (capacity) of the hypothesis class that can be learned by a statistical classification algorithm. First existing bounds for simple learning algorithms, such as the histogram classifier, computed the complexity as the cardinality of the hypothesis class~\cite{ruicastrolecture1}. The corresponding bounds were inapplicable to problems involving an infinite class of functions, for which they became very loose. This led to the development of a new capacity measure - the VC dimension~\cite{Vapnik:1982:EDB:1098680,Vapnik2015,vapnik2013nature,sauer1972density,svm,ehrenfeucht1989general}, which is defined as the cardinality of the largest set of points that the algorithm can shatter and thus does not scale with the size of the hypothesis class. Resolving the VC bounds for neural networks~\cite{JMLR:v20:17-612,anthony_bartlett_1999} leads to the theoretical guarantees that depend on the number of network parameters. Such bounds are not useful for practical networks.

The aforementioned error bounds are distribution-free and often loose in practice. This motivated work on distribution-dependent capacity measures such as VC entropy~\cite{Vapnik1998}, covering numbers~\cite{alon1997scale,zhang2002covering}, and Rademacher complexity~\cite{koltchinskii2001rademacher,bartlett2002rademacher, luxburg2004distance}. Bounds based on covering numbers were derived for a limited family of classifiers, such as linear functional classes or neural networks with identity activation functions~\cite{zhang2002covering}. Rademacher complexity measures the ability of functions in the hypothesis space to fit to random labels. It has been recently observed that DNNs are powerful enough to fit any set of random labels~\cite{DBLP:conf/iclr/ZhangBHRV17} thus rendering the Rademacher complexity based bounds inadequate. Other capacity measures for neural networks, not mentioned before, include  unit-wise capacities~\cite{DBLP:journals/corr/abs-1805-12076}. They led to generalization bounds for two layer ReLU networks. An excellent comparison of existing  DNN generalization measures can be found in~\cite{Fan2020}.

Estimating generalization bounds using PAC-Bayesian approaches and margin based analysis ~\cite{mcallester1999pac,mcallester2003simplified,langford2003pac} is still an active area of research. More recently, several bounds based on PAC-Bayes approach have been presented for stochastic and compressed networks~\cite{dziugaite2017computing, arora2018stronger, zhou2018non} that are computational in nature, by exploring modifications of the standard training procedure to obtain tighter(non-vacuous) generalization guarantees. However, these bounds are still loose ($>>0$) to practically study the sample complexity of DNNs.


There also exist works that study the generalization phenomenon in DL from the perspective of the behavior of the optimization algorithm that minimizes the training loss. They are focused on the convergence properties of the optimizers and therefore are outside of the focus of this paper, with the exception of ~\cite{neyshabur2014search} that argues the existence of the ``inductive bias'' imposed by the optimizer, such as SGD, that restricts neural networks to a simple class of functions. This idea is linked with the notion of network's capacity though it is unclear how to use it to obtain sample complexity guarantees.

Above we discussed works that aim at proving theoretical bounds on the generalization error. Existing bounds that hold only for simplified DNNs typically scale with the training data size as $\mathcal{O}(1/\sqrt{N})$. An empirical family of approaches, that we will discuss next, instead studies the ways of extrapolating the learning curves, i.e. the dependence of the error on the amount of the training data, using parametric models. Among these works, we have linear, logarithmic, exponential, and power law parametric models~\cite{DBLP:conf/aistats/FreyF99} that were applied to decision trees. A subsequent paper~\cite{gu2001modelling} explored a vapor pressure model, the Morgan Mercer-Flodin (MMF) model, and the Weibull model to predict learning curves for classification algorithms such as decision tree and logistic discrimination. Empirical results obtained for a $2$-layer neural network on MNIST data set showed that learning curve decays following the power law with a decay factor in the range $[1,2]$~\cite{cortes1994learning,cortes1995limits}. This behavior of the learning curve was also observed in other applications~\cite{DBLP:journals/corr/abs-1712-00409}. These parametric modeling approaches are not supported by theoretical argument.

Finally, research works that are most closely related to our approach present asymptotic estimates of learning curves for Gaussian processes~\cite{sollich2002gaussian,williams2000upper}, kernel methods~\cite{spigler2019asymptotic}, and wide neural networks trained in the regime of neural tangent kernels~\cite{cohen2019learning}. These works do not apply to a practical deep learning setting, but provide useful insights into the mathematical modeling of complex learning phenomena. 

\section{Generalization error estimation}
\label{sec:theoretical_framework}
In this section we derive the estimates for the generalization error of a DL model. Our analysis is performed under the assumptions that the model can learn the training data set with perfect accuracy and the probability of making an error on the test examples takes the form given in Equation~\ref{eq:phi_of_x}.

\subsection{Effective dimensionality}
Practical DNNs are over-parameterized, i.e., the number of parameters far exceeds the number of training data samples. This over-parameterization induces redundancy in network weights~\cite{denton2014exploiting,DBLP:conf/iclr/MolchanovTKAK17,han2015learning}, which particularly manifests itself on the output of the feature extractor of the network (the feature extractor typically precedes the fully connected layers of the model). The data representation there has to be simple enough so that the last layers of the network, which constitute shallow classifier, can perform accurate prediction. It has been noted in past works that this feature vector is low dimensional~\cite{ravichandran2019using,DBLP:journals/corr/abs-1804-07090,DBLP:journals/corr/abs-1906-00443,DBLP:journals/corr/abs-1905-12784}. We next describe how we define and find effective dimensionality of the feature space.

We introduce a bottleneck network consisting of two linear layers, each followed by the ReLU non-linearity, before the output layer of the network. The bottleneck takes $D$-dimensional feature vector as input, projects it down to dimensionality $d^{'}$, and then projects it back up to input dimension $D$. We insert the bottleneck into the trained model and fine-tune the entire network. The effective dimensionality $d$ is the smallest value of $d^{'}$, for which the accuracy of the model with the bottleneck does not differ significantly from the accuracy of the original model without it. Empirical evaluation of $d$ for different networks and data sets is presented in the experimental section.

The existence of small effective dimensionality of the feature space has been observed before in various works. Specifically,~\cite{ravichandran2019using} defines effective dimensionality of the feature maps in terms of singular values of its co-variance matrix. They observe that as we move from input to the output layer of the network the effective dimensionality first increases and then drops. They report an effective dimensionality at the final layer of the network and show that it is as low as 2 for tiny ImageNet and CIFAR-10 data sets. Furthermore, they also observe a much sharper decline in effective dimensionality for large networks compared to the small ones. Similarly,~\cite{DBLP:journals/corr/abs-1804-07090} observed that $<10$ singular values of the matrix of vectorized representations are enough to explain $>99\%$ of the variance. They noted that enforcing even stronger low rank structure for the feature co-variance matrix can lead to better performance and robustness to adversarial examples.~\cite{DBLP:journals/corr/abs-1906-00443,DBLP:journals/corr/abs-1905-12784} utilize an ``ID estimator" previously introduced in~\cite{facco2017estimating} that relies on the ratio of distances to the nearest and second nearest neighbor of a data point to analyze intrinsic network dimensionality. These authors also observe that the neural network first increases and then decreases its intrinsic dimensionality to as low as $10$ when moving towards network's output. Another work~\cite{goldfeld2018estimating} reports similar behavior of the mutual information. The mutual information was found to be as low as $<4$ nats closer to the final layer of the neural network. Finally, numerous network compression approaches implicitly rely on the existence of small effective dimensionality of the feature space when pruning network connections. They achieve $\approx 90\%$~\cite{DBLP:conf/iclr/ZhuG18} compression rate with negligible loss of the accuracy of the model.

We next move to our mathematical modeling of the generalization error. For the purpose of simplifying our analysis, we first consider the case where the effective dimensionality of the feature space is one and then extend the analysis to the general case of arbitrary dimensionality. Let $f_{train}$ and $f_{test}$ denote probability density functions of the train and test feature distributions. Then under the proposed error model defined in Equation~\ref{eq:phi_of_x} the overall probability of making an error on the test set is given by the expectation $\mathbb{E}_{f_{\text{test}}}[\Phi]$.

\vspace{-5pt}
\subsection{Generalization error estimates for one dimensional case ($d$ = 1)}
Let $\hat{x}$ be a given test point in the feature space whose immediate nearest training points in the feature space are $x_{i}$ and $x_{j}$, such that $\hat{x} \in (x_{i}, x_{j})$ and let $\rho(\hat{x}) = |x_{j} - x_{i}|$. Intuitively, as we increase the number of training data points sampled from $f_{train}$ the distance between two training samples i.e $\rho(x)$ decreases. Assume that $f_{test}$ is close to a uniform distribution, denoted as $u$, in the interval $(x_{i}, x_{j})$. Note that this is a realistic assumption, i.e. at the tail of the distribution we observe training samples rarely but at the same time the training distribution there is flat whereas in high-concentration regions, where the training distribution changes quickly, the training samples are observed close to each other. Thus in the latter case the dynamics of the changes of the distribution are compensated by the small distance between samples. Also the more data we have, which is the regime we are mostly interested in analyzing, the more accurate this assumption is.
Since the test point $\hat{x}$ is uniformly distributed in the interval $(x_{i}, x_{j})$, the distance from the test point to its closest training point (denoted as $\psi(\hat{x})$) is also uniformly distributed in the range $[0, \frac{\rho(\hat{x})}{2}]$. We can compute the expectation of $\psi(\hat{x})$ as (see Derivations for Equation~\ref{eq:exp_psi_1d} in the Supplement),
\begin{equation}
   \label{eq:exp_psi_1d}
   \mathbb{E}_u^{\langle x_{i}, x_{j} \rangle}[\psi(\hat{x})] = \frac{|x_j - x_i|}{4} = \frac{\rho(\hat{x})}{4}.
\end{equation}
In the large data regime, we can approximate the distance between two training points ($\rho(\hat{x})$) as the limit of the ratio of length of the interval to number of points lying in the interval as,
\begin{align*}
    & \rho(\hat{x}) \approx  \lim_{\Delta \to 0} \frac{\Delta}{\int_{\hat{x} - \frac{\Delta}{2}}^{\hat{x} + \frac{\Delta}{2}}Nf_{\text{train}}(x)dx} \\
    &= \lim_{\Delta\to 0}\frac{\Delta}{N[F_{\text{train}}(\hat{x} + \frac{\Delta}{2}) - F_{\text{train}}(\hat{x} -\frac{\Delta}{2})]} = \frac{1}{Nf_{\text{train}}(\hat{x})}
\end{align*}

The above approximation does not include the local variance of $\rho(\hat{x})$. The effect of local variance results from the fact that neighboring training intervals should roughly have the same length but in practice they do not. Including that effect is crucial in the experiments. Thus we correct $\rho(\hat{x})$ by taking into account this local variance. We denote corrected $\rho(\hat{x})$ as $\rho'(\hat{x})$. The neighboring intervals should have same density function usually, so we calculate $\rho'(\hat{x})$ using $K$ left and $K$ right neighboring intervals of the training interval $\langle x_{i},x_{j}\rangle$. We refer to the lengths of these intervals as $\rho_{-K}(\hat{x}), \rho_{-K+1}(\hat{x}), \dots, \rho_{K}(\hat{x})$. 
Note that 
\begin{align*}
    \mathbb{E}^{\langle \rho_{-K},\rho_K\rangle}[\rho(\hat{x})]\coloneqq \frac{\sum_{i=-K}^K\rho_i(\hat{x})}{2K+1} = \rho(\hat{x})
\end{align*}
and 
\begin{align*}
    \text{Var}^{\langle \rho_{-K},\rho_K\rangle}[\rho(\hat{x})] \coloneqq \frac{\sum_{i=-K}^K\rho_i^2(\hat{x})}{2K+1} - \left(\mathbb{E}^{\langle \rho_{-K},\rho_K\rangle}[\rho(\hat{x})]\right)^2
\end{align*}

\begin{align*}
    & \rho'(\hat{x}) = \sum_{i=-K}^K\underbrace{\frac{\rho_i(\hat{x})}{\sum_{j=1}^K\rho_j(\hat{x})}}_{\text{prob. of falling into the interval}}\cdot\underbrace{\rho_i(\hat{x})}_{\text{interval length}}\\
    &= \frac{1}{\sum_{j=-K}^K\rho_j(x_t)}\sum_{i=-K}^K\rho_i(\hat{x})^2dx = \frac{\sum_{i=-K}^K\rho_i(\hat{x})^2}{\sum_{j=-K}^K\rho_j(\hat{x})} \\
    &= \frac{\text{Var}^{\langle \rho_{-K},\rho_K\rangle}[\rho(\hat{x})]}{\mathbb{E}^{\langle \rho_{-K},\rho_K\rangle}[\rho(\hat{x})]} +\mathbb{E}^{\langle \rho_{-K},\rho_K\rangle}[\rho(\hat{x})]
\end{align*}

\noindent For $1$-dimension case, we empirically verified $\mathbb{E}^{\langle \rho_{-K},\rho_K\rangle}[\rho(\hat{x})] \approx \frac{Var^{\langle \rho_{-K},\rho_K\rangle}[\rho(\hat{x})]}{\mathbb{E}^{\langle \rho_{-K},\rho_K\rangle}[\rho(\hat{x})]}$, thus:

\begin{align}
    \label{eq:exp_rho_1d}
    \rho'(\hat{x}) = 2 \rho(\hat{x}) = \frac{2}{Nf_{\text{train}}(\hat{x})}
\end{align} 

Using Equations~\ref{eq:phi_of_x},~\ref{eq:exp_psi_1d}, and~\ref{eq:exp_rho_1d} we can derive the probability of an error on the test set, $\mathbb{E}_{f_{\text{test}}}[\Phi]$, as follows:

\begin{align}
    &\mathbb{E}_{f_{\text{test}}}[\Phi] = \int_{-\infty}^{+\infty} \Phi(\hat{x})f_{\text{test}}(\hat{x})d\hat{x} \nonumber\\ 
    &= \int_{-\infty}^{+\infty} min\left(1, \frac{\psi(\hat{x})}{\delta}\right)f_{\text{test}}(\hat{x})d\hat{x}\nonumber\\
    &\approx \int_{-\infty}^{+\infty} min\left(1, \frac{\rho(\hat{x})}{4\delta}\right)f_{\text{test}}(\hat{x})d\hat{x} \nonumber\\
    &\approx \int_{-\infty}^{+\infty} min\left(1, \frac{1}{2Nf_{\text{train}}(\hat{x})\delta}\right)f_{\text{test}}(\hat{x})d\hat{x} \qedhere
    \label{eq:exp_phi_test_1d}
\end{align}

The integral in Equation~\ref{eq:exp_phi_test_1d} can be computed in the closed form for many standard distributions, such as Gaussian or uniform, else it can be computed using Monte Carlo method~\cite{caflisch_1998}.

\subsection{Generalization error estimates for multi-dimensional case}
\begin{figure*}[ht]
    \centering
    \includegraphics[width=0.32\textwidth]{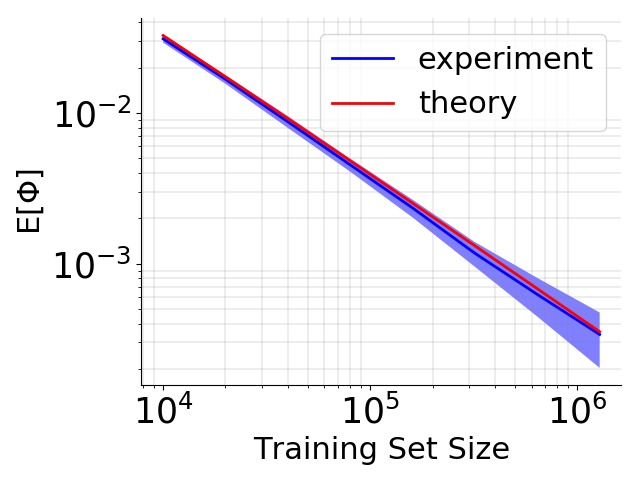}
    \includegraphics[width=0.32\textwidth]{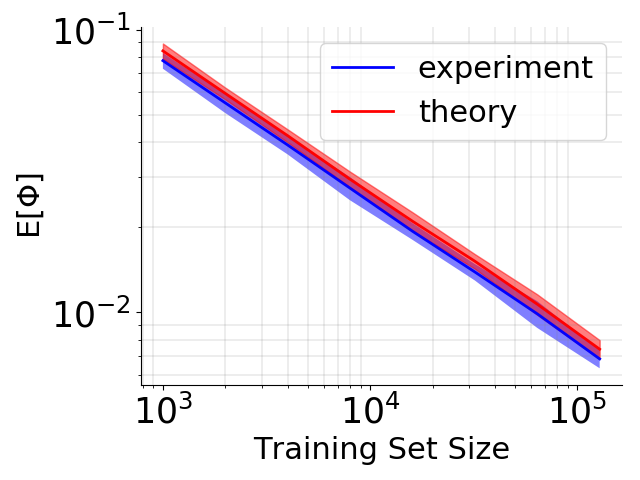}
    \includegraphics[width=0.32\textwidth]{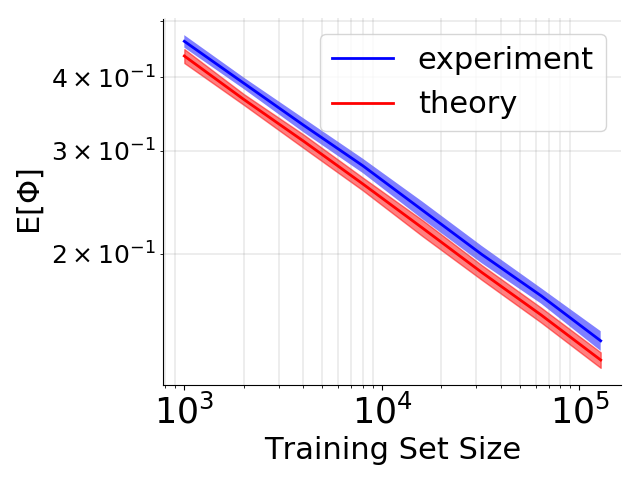}\\
    \caption{Monte Carlo simulation results (blue curve) confronted with theoretical derivations (red curve) for \textbf{(left)} d = 1 \textbf{(center)} d=2 and \textbf{(right)} d = 4. M = 1K, $f_{train}$ = $f_{test}$ = $\mathcal{N}_d(\mu=0, \Sigma=I)$. The error bars capture $2$ standard deviations.}
    \label{fig:brute_force_results}
    \vspace{-15pt}
\end{figure*}

Now we consider multi-dimensional feature distributions. Let $\hat{x}$ be a test point in the feature space whose immediate $2^d$ nearest training points in the feature space form a set $\bar{X}$ and let $\mathcal{P}$ be a convex hull spanned by these training points. Assume $\mathcal{P}$ contains $\hat{x}$. For the ease of further derivations, we assume training points from $\bar{X}$, sampled from distribution $f_{train}$, lie on the vertices of a $d$-dimensional hyper-cube $\mathcal{P}$ with side length $a(\hat{x})$. The side length of the hyper-cube $\mathcal{P}$ depends on the position of the test point $\hat{x}$. This is because in places with higher density of training data points we can construct a tighter convex hull around the test point $\hat{x}$, and hence the length of the side of the hyper-cube $\mathcal{P}$ should decrease then. Furthermore, similar to 1-dimensional case let the test feature distribution be close to uniform, denoted as $u$, in $\mathcal{P}$. 

In the large data regime, we approximate the distance of $\hat{x}$ to its closest training feature vector (denoted as $\psi(\hat{x})$) with the expected value of the distance of $\hat{x}$ to the closest training point in $\mathcal{P}$ (depending on the position in $\mathcal{P}$, the closest training point is one of the vertices of the hyper-cube $\mathcal{P}$). For ease of computation we assume $x(\hat{x})$ lies at the origin of the $d$-dimensional feature space hence, we can compute $\mathbb{E}_u^{\bar{X}}[\psi(\hat{x})]$ as,

\begin{align}
    \label{eq:psi_hd}
    &\mathbb{E}_u^{\bar{X}}[\psi(\hat{x})] = \int_{\mathcal{P}}\|\hat{x}-x(\hat{x})\|_2u(\hat{x})d\hat{x} \nonumber \\
    &= \int_0^\frac{a(\hat{x})}{2}\dots\int_0^\frac{a(\hat{x})}{2}\|\hat{x}\|_2\frac{1}{(\frac{a(\hat{x})}{2})^d}d\hat{x}_1\dots d\hat{x}_d \nonumber \\
    &= \frac{1}{(\frac{a(\hat{x})}{2})^d}\int_0^\frac{a(\hat{x})}{2}\dots\int_0^\frac{a(\hat{x})}{2}\sqrt{\sum_{i=1}^d\hat{x}_i^2}d\hat{x}_1\dots d\hat{x}_d
\end{align}

In the large data regime, we can approximate the distance between two training data points, or in other words the side of the hyper-cube $\mathcal{P}$, $a(\hat{x})$, as the limit of the ratio of the volume of the hyper-cube $\mathcal{P}$ to the number of points lying in $\mathcal{P}$: 
\color{black}
\begin{align}
    \label{eq:a_hd}
    a(\hat{x}) &\approx \left(\lim_{\text{Volume}(\mathcal{P}) \to 0} \frac{\text{Volume}(\mathcal{P})} {\int_{\mathcal{P}}Nf_{\text{train}}(\hat{x})dx}\right)^{1/d} \nonumber \\
    &= \left(\lim_{\Delta\to 0} \frac{1}{Nf_{\text{train}}(\hat{x} + \Delta)}\right)^{1/d} \nonumber \\
    &= \frac{1}{(Nf_{\text{train}}(\hat{x}))^{1/d}}.
\end{align}
In higher dimensions ($>$ 1), we empirically verified that no correction to $a(\hat{x})$ is required.

Similarly to 1-dimensional case, we can use Equations~\ref{eq:phi_of_x},~\ref{eq:psi_hd}, and~\ref{eq:a_hd} to derive the probability of an error on the test set, $\mathbb{E}_{f_{\text{test}}}[\Phi]$, as follows:
\begin{align}
\label{eqn:Exp_Phi_test_hd}
\mathbb{E}_{f_{\text{test}}}[\Phi] =& \int\limits_{-\infty}^{+\infty}\!\!\dots\!\!\int\limits_{-\infty}^{+\infty}\min\left(1, \frac{\psi(\hat{x})}{\delta}\right)f_{\text{test}}(\hat{x})d\hat{x}
\end{align}
where, 
\vspace{-0.7cm}
\begin{align*}
    \psi(\hat{x}) = & \frac{1}{(\frac{a(\hat{x})}{2})^d} \int\limits_0^\frac{a(\hat{x})}{2}\!\!\dots\!\!\int\limits_0^\frac{a(\hat{x})}{2}||\hat{x}||_{2}d\hat{x}_1 \dots d\hat{x}_d
\end{align*}
and
\vspace{-0.7cm}
\begin{align*}
    a(\hat{x}) =& \frac{1}{(Nf_{train}(\hat{x}))^{\frac{1}{d}}}.
\end{align*}

The obtained integral cannot be computed in the closed form, however it can be computed using Monte Carlo methods (note that $d$ in our experiments is very small, i.e. it does not exceed $4$, which enables accurate Monte Carlo approximations).

\section{Experiments}\label{sec:experiments}
We conduct two types of experiments. First, we verify our derivations for the generalization error estimator using Monte Carlo simulations. We use toy data sets generated from the Gaussian distribution. We then move to the main experiments, which are performed on the real data. These experiments involve classification and regression problems. The classification task is performed on the following data sets: MNIST~\cite{lecun-mnisthandwrittendigit-2010}, CIFAR~\cite{cifar} and ImageNet~\cite{imagenet_cvpr09}). Our experiments utilize popular DNN architectures: LeNet~\cite{lecun-mnisthandwrittendigit-2010}, VGG16~\cite{simonyan2014very}, ResNet18, and ResNet50~\cite{DBLP:journals/corr/HeZRS15}.  We used cross entropy loss functions and stochastic gradient descent at training. The regression task is performed on the Udacity~\cite{udacitydata} data set, which is typically used in the autonomous driving applications. It contains images from left, center and right cameras that are mounted on the vehicle and additional vehicle logs such as speed, steering command etc. The data set is imbalanced and contains mostly samples corresponding to driving straight. We sub-sampled those to balance the data. The final balanced data set contains $38936$ training examples, $6552$ validation examples, and $8190$ test examples. For the Udacity experiments we utilize a network described in Table~\ref{tab:CovNet} in the Supplement that takes single image as input and predicts the appropriate steering command. The network was trained using mean squared error loss and Adam optimizer~\cite{DBLP:journals/corr/KingmaB14}. 
\begin{figure*}[!ht]
    \centering
    \includegraphics[width=0.32\textwidth]{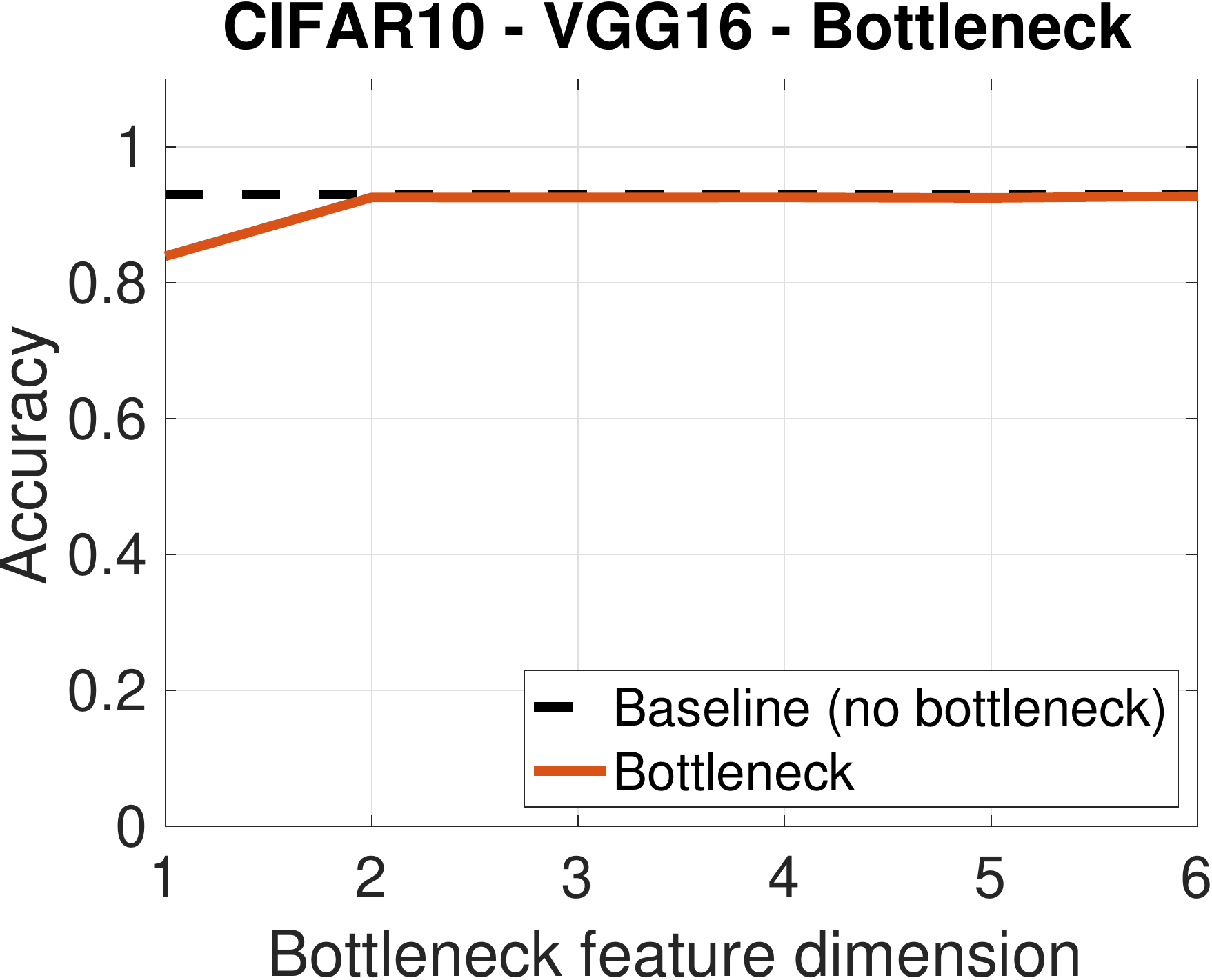} 
    \includegraphics[width=0.32\textwidth]{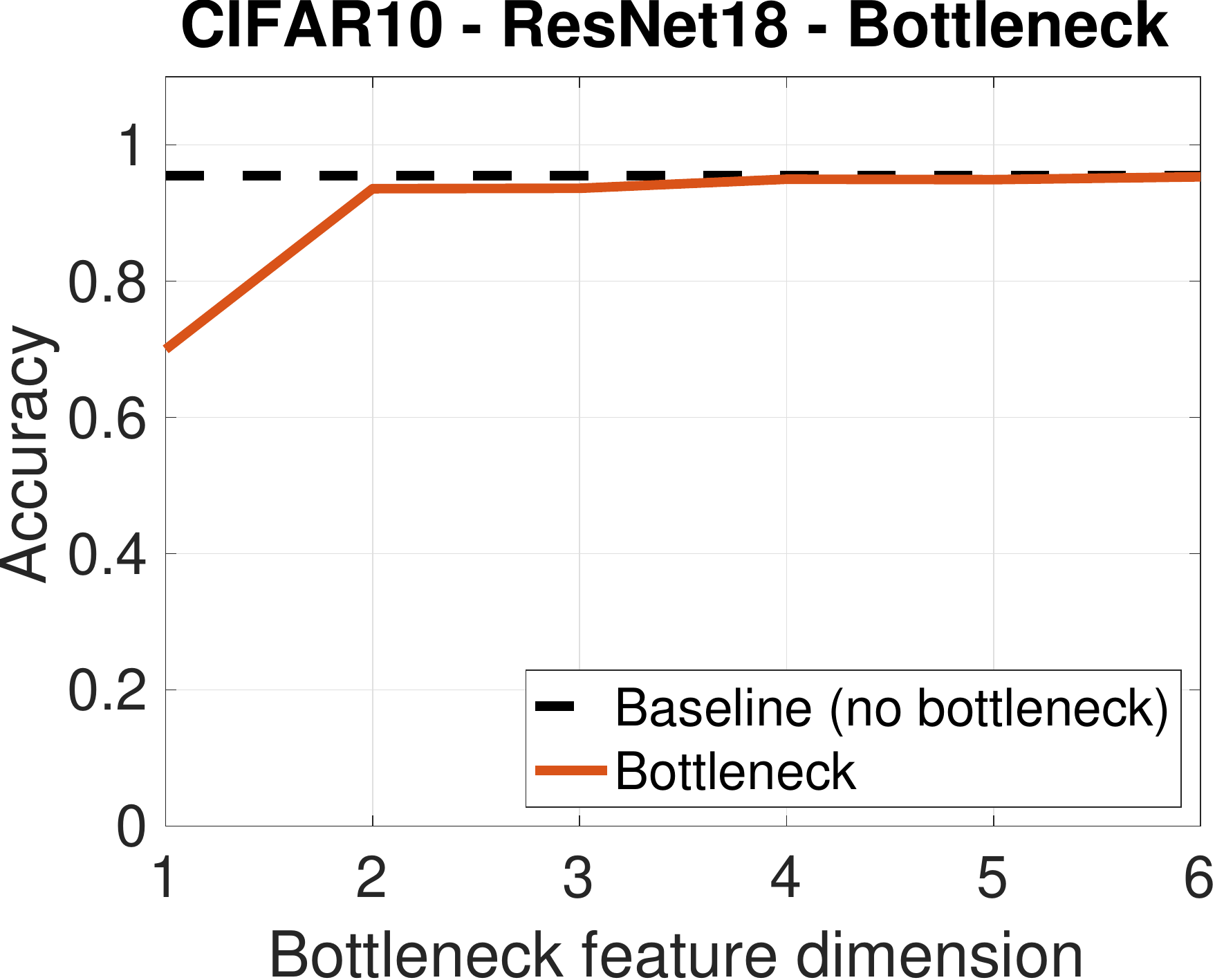}
    \includegraphics[width=0.32\textwidth]{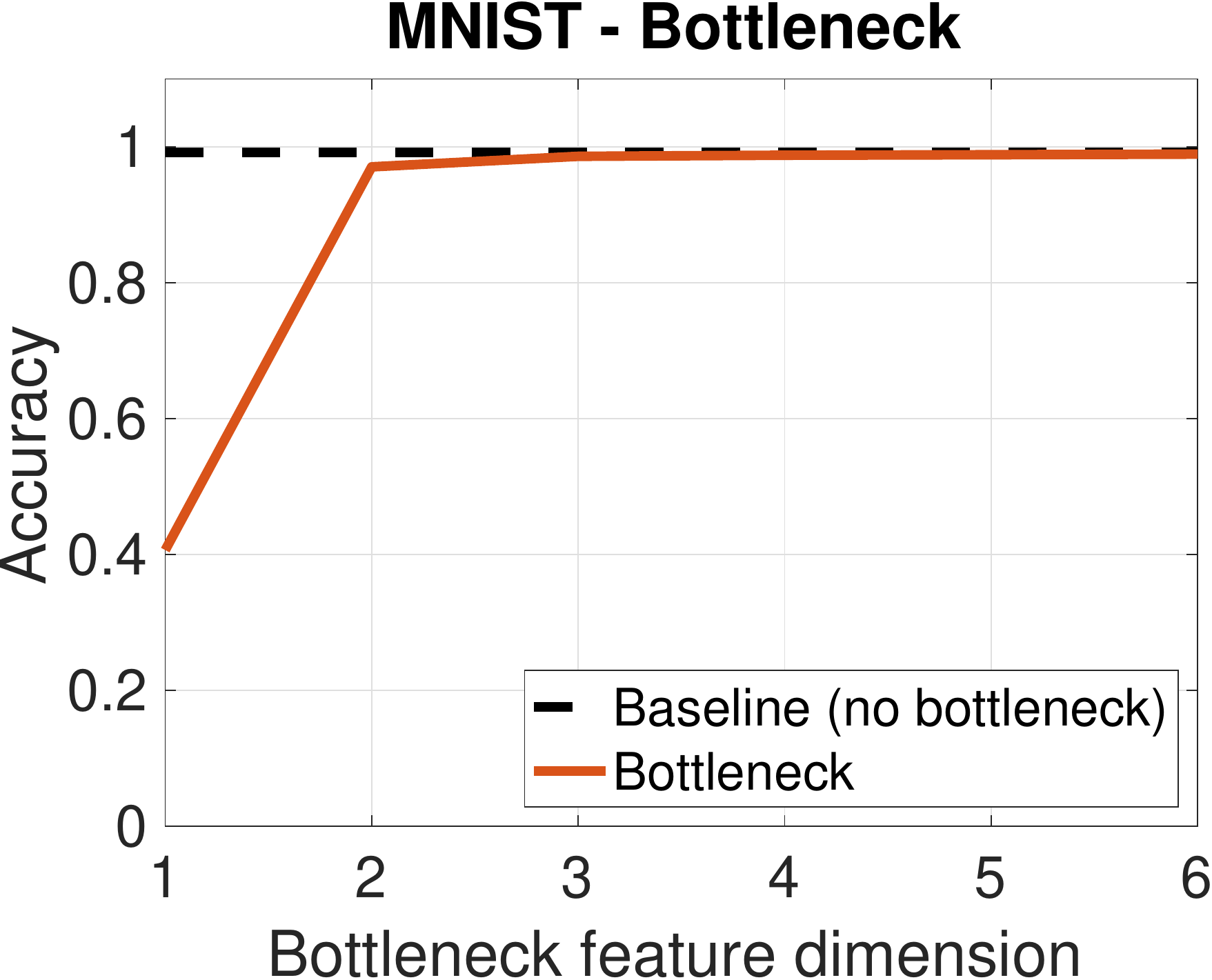} \\
    \vspace{5pt}
    \includegraphics[width=0.32\textwidth]{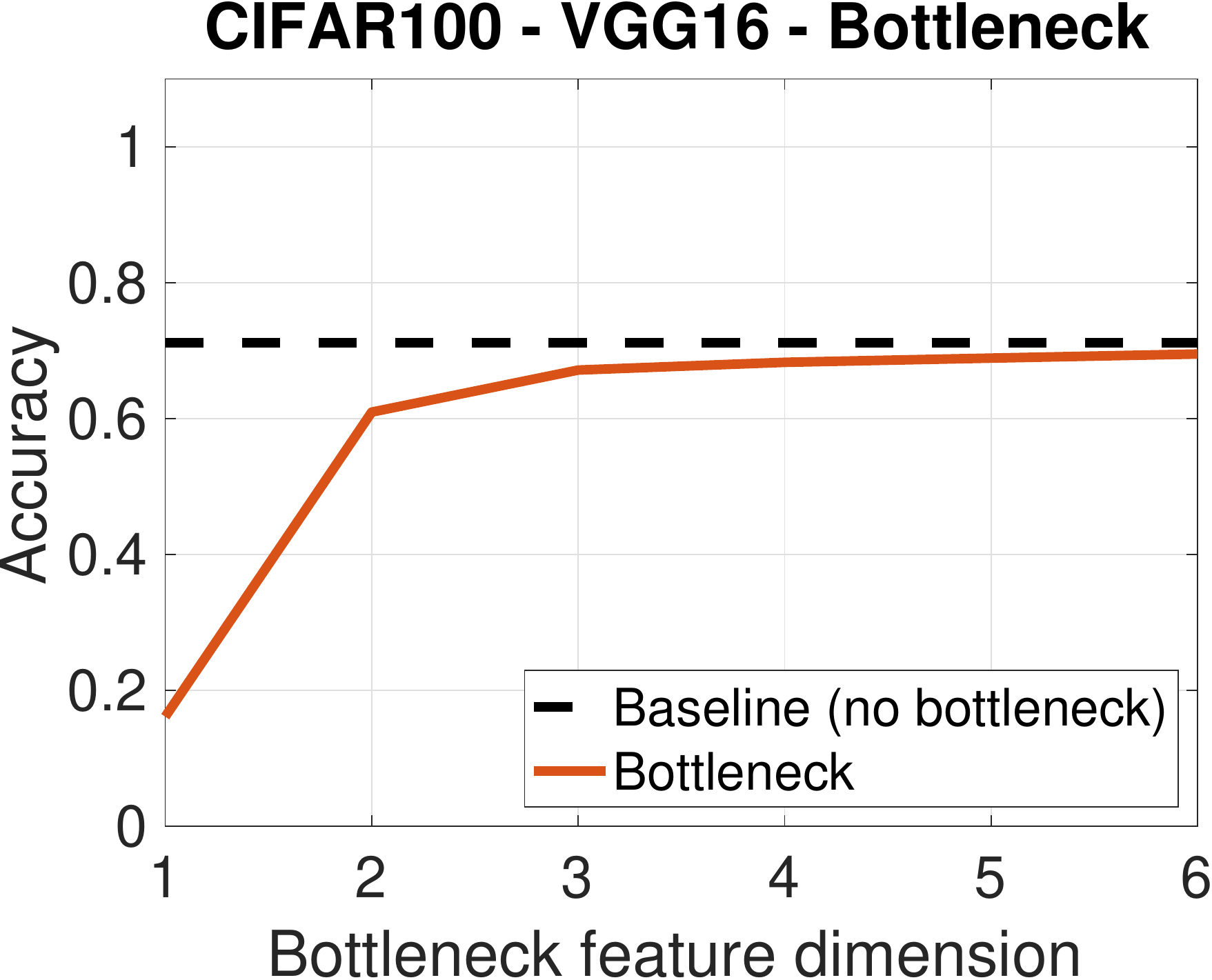}
    \includegraphics[width=0.32\textwidth]{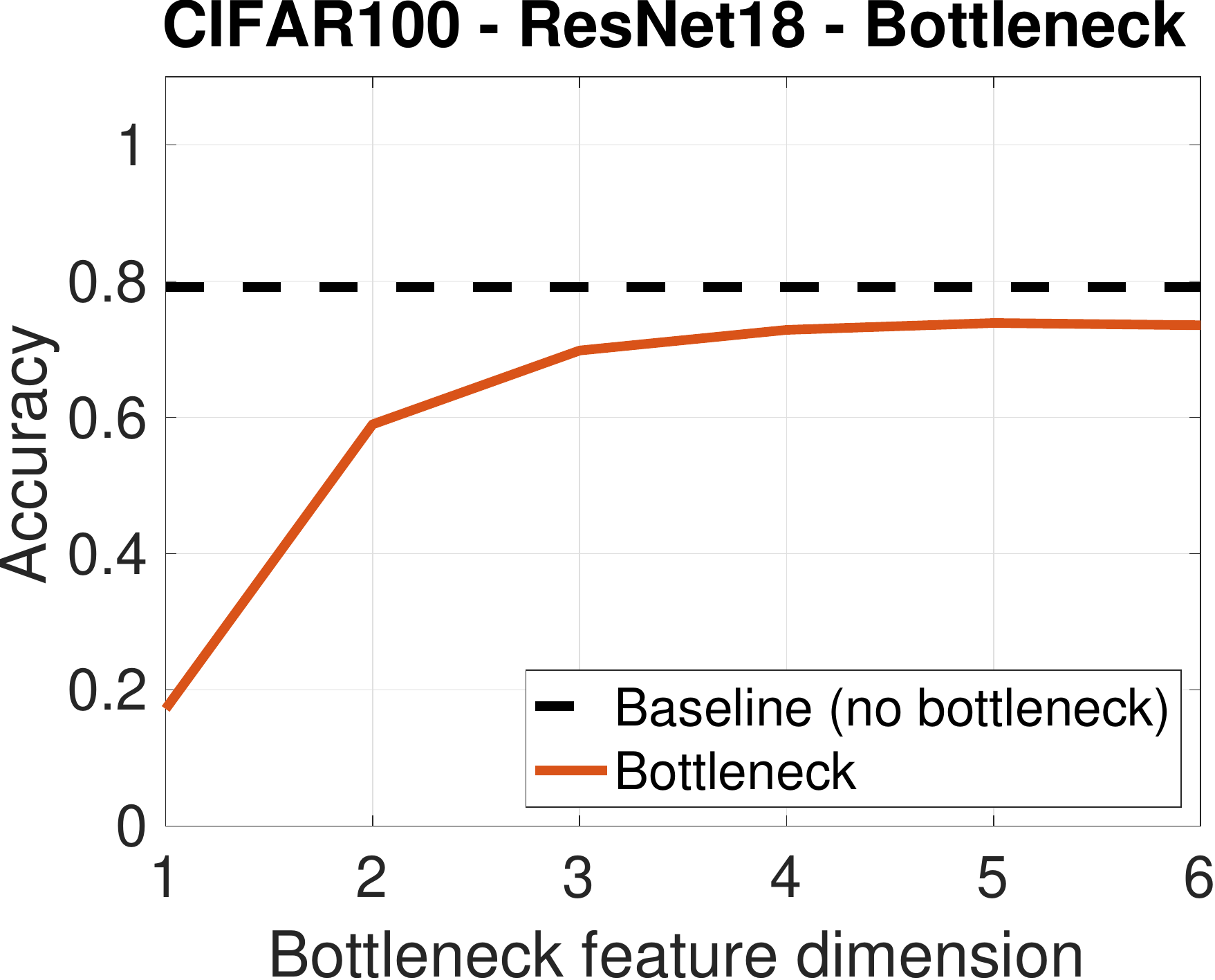}
    \includegraphics[width=0.32\textwidth]{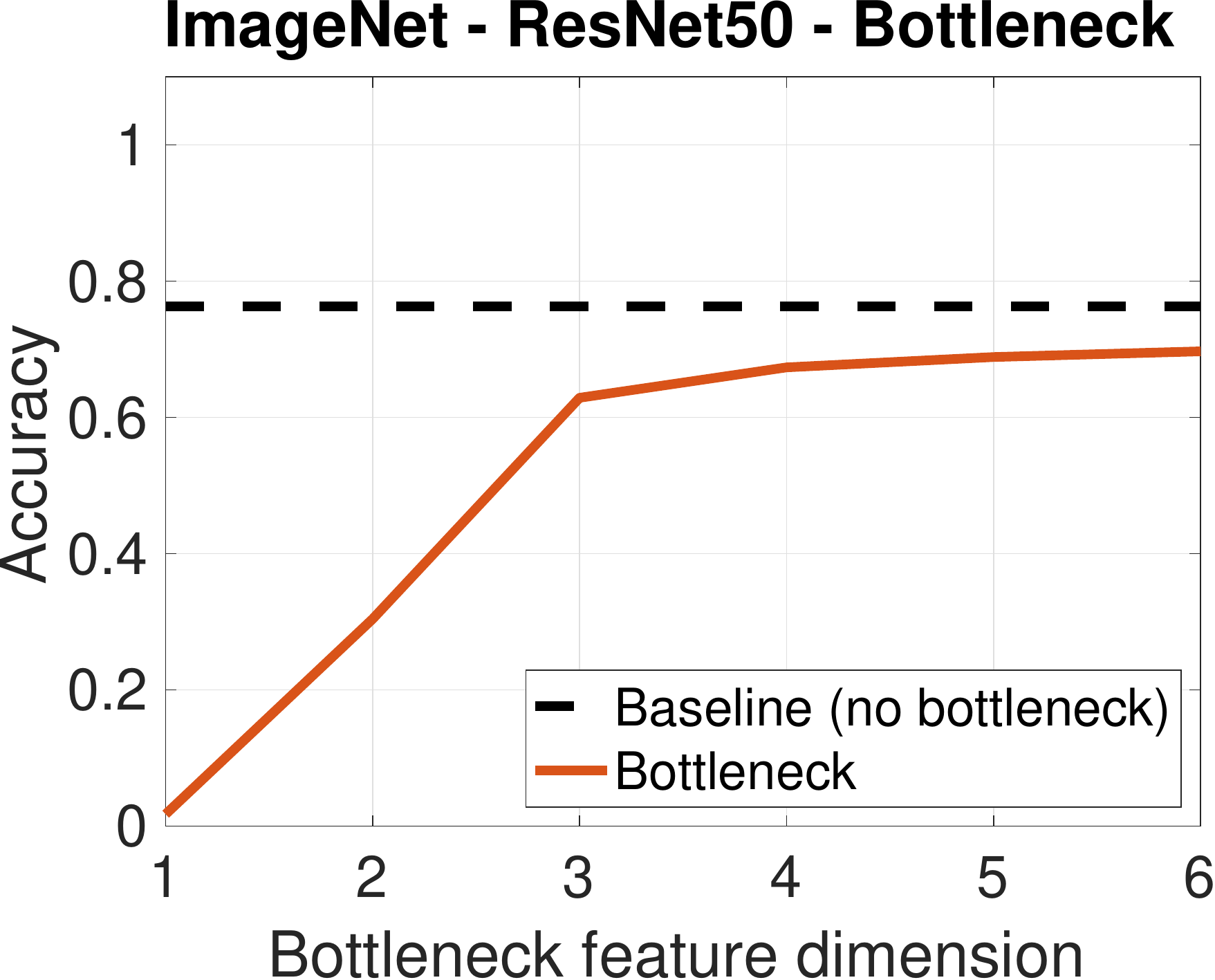}
    \caption{Test accuracy for different classification data sets with bottleneck network inserted before the output layer of the model. \textbf{Top Row:} CIFAR-10 trained on \textbf{(Left)} VGG16, \textbf{(Center)} ResNet-18 and \textbf{(Right)} MNIST trained on LeNet.  \textbf{Bottom Row:} CIFAR-100 trained on \textbf{(Left)} VGG16 , \textbf{(Center)} ResNet-18 and \textbf{(Right)} ImageNet trained on ResNet-50.}
    \label{fig:bottleneck}
    \vspace{-15pt}
\end{figure*}
\subsection{Monte Carlo simulations}\label{sec:bfexp}
Here we assume that both train and test data points are $d$-dimensional vectors (we explored $d=1,2,4)$ that are drawn from a known Gaussian distribution $\mathcal{N}_d(\mu, \Sigma)$. We set $\delta = 1$. We generate the train and test sets containing respectively $N$ and $M$ data points. Since we are interested in this paper in examining how the error scales with $N$, our experiments are performed on training data sets with a growing size ($N$). In the simulation, for each test point we find the closest point in the training data set. We count the test point as a failure with the probability obtained using Equation~\ref{eq:phi_of_x}. The error rate is computed as a number of failures divided by the size of the test data set (blue curve in Figure~\ref{fig:brute_force_results}). We run the simulation for each value of $N$ twenty time with different seeds. We confront the error rate obtained from simulation with the theoretical one obtained using Equations ~\ref{eq:exp_phi_test_1d} and ~\ref{eqn:Exp_Phi_test_hd} (red curve in Figure~\ref{fig:brute_force_results}). We use Monte Carlo method to compute the integrals in these equations.
The results are captured in Figure~\ref{fig:brute_force_results}. The experiment shows that simulated and theoretical curves match, which confirms the correctness of our theoretical derivations. 

\color{black}
\subsection{Real data experiments} \label{sec:exps_main}
\subsubsection{Finding effective dimensionality}
\label{sec:d_eff}
\begin{figure*}[!t]
    \centering
    \includegraphics[width=0.32\textwidth]{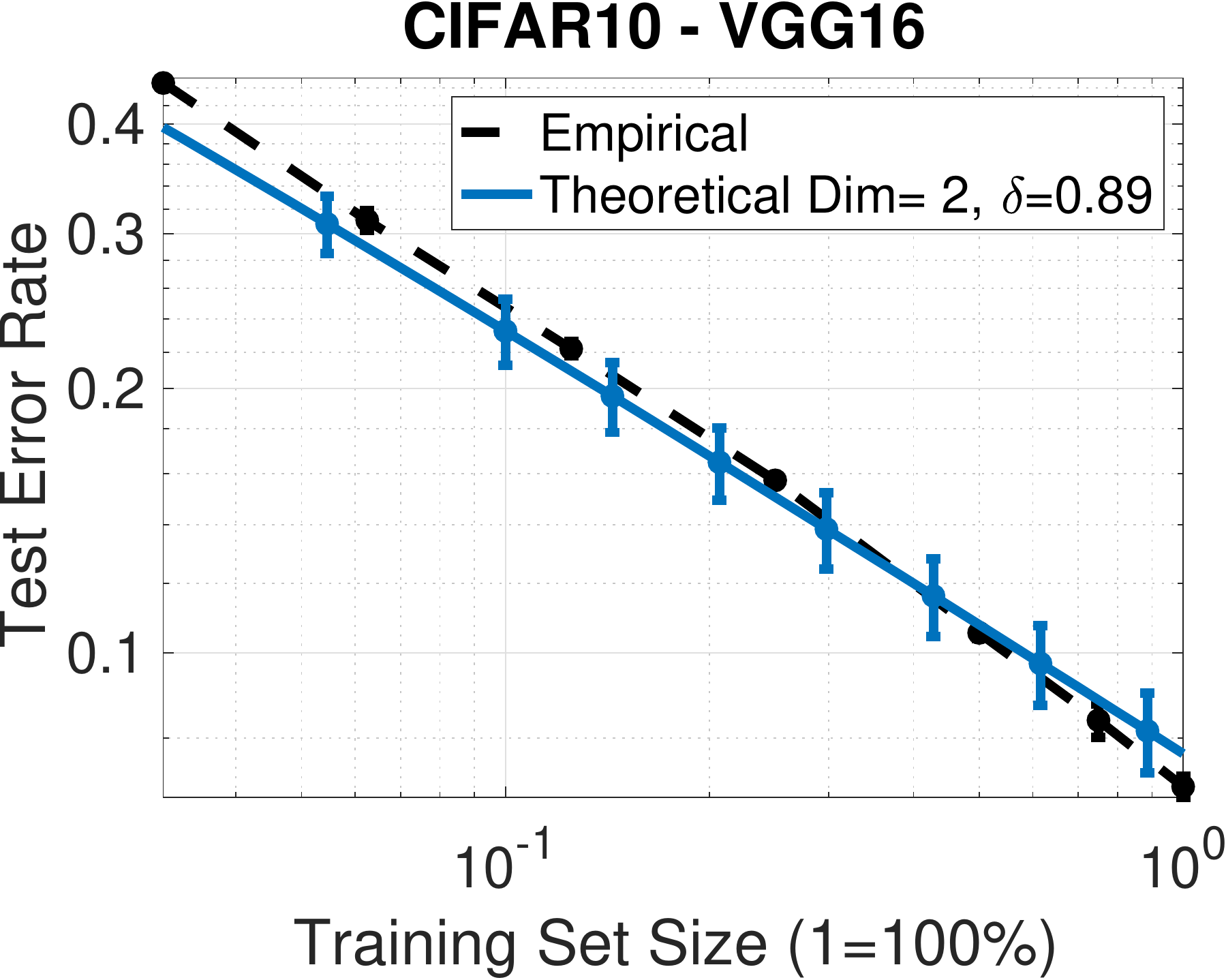}
    \includegraphics[width=0.32\textwidth]{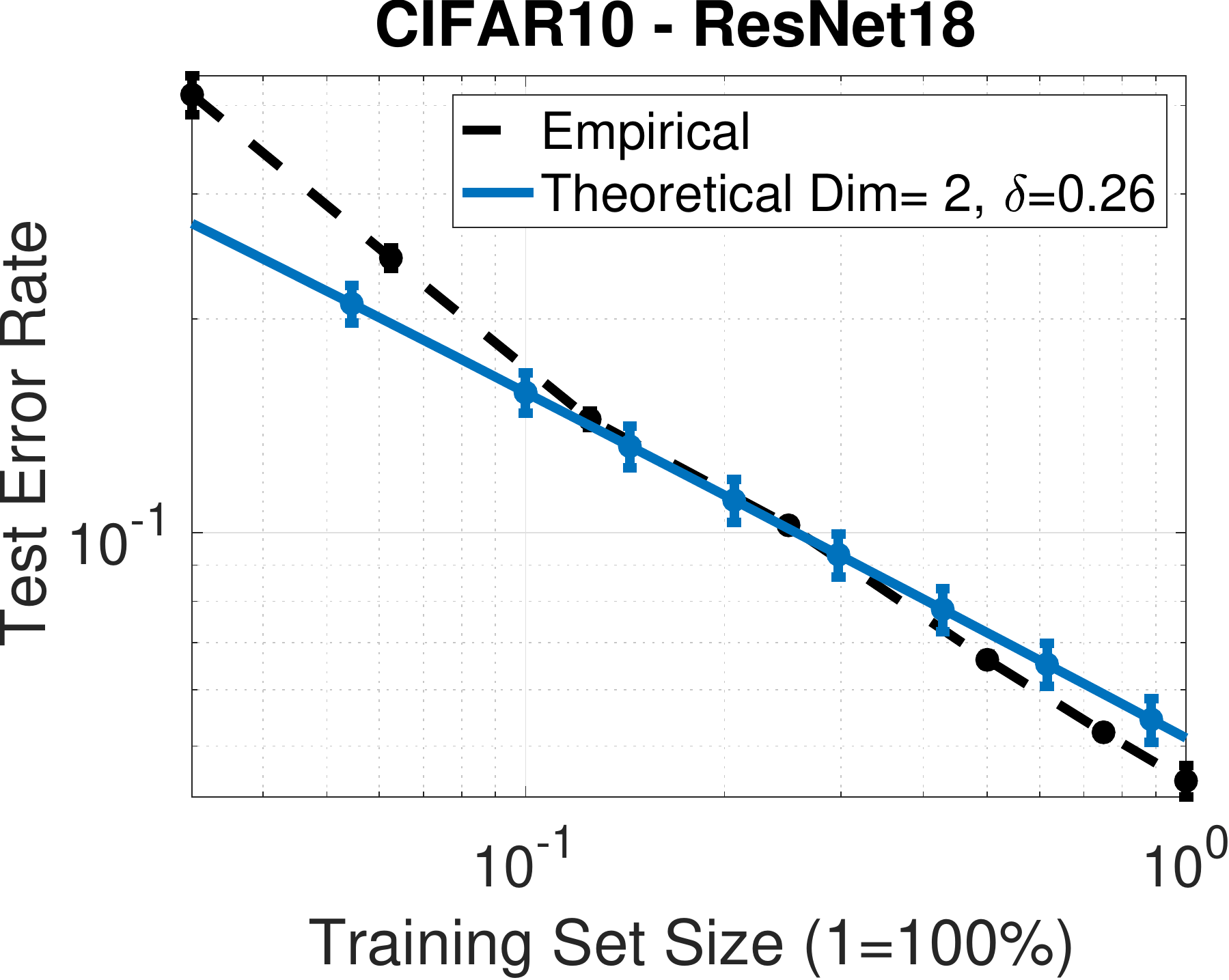}
    \includegraphics[width=0.32\textwidth]{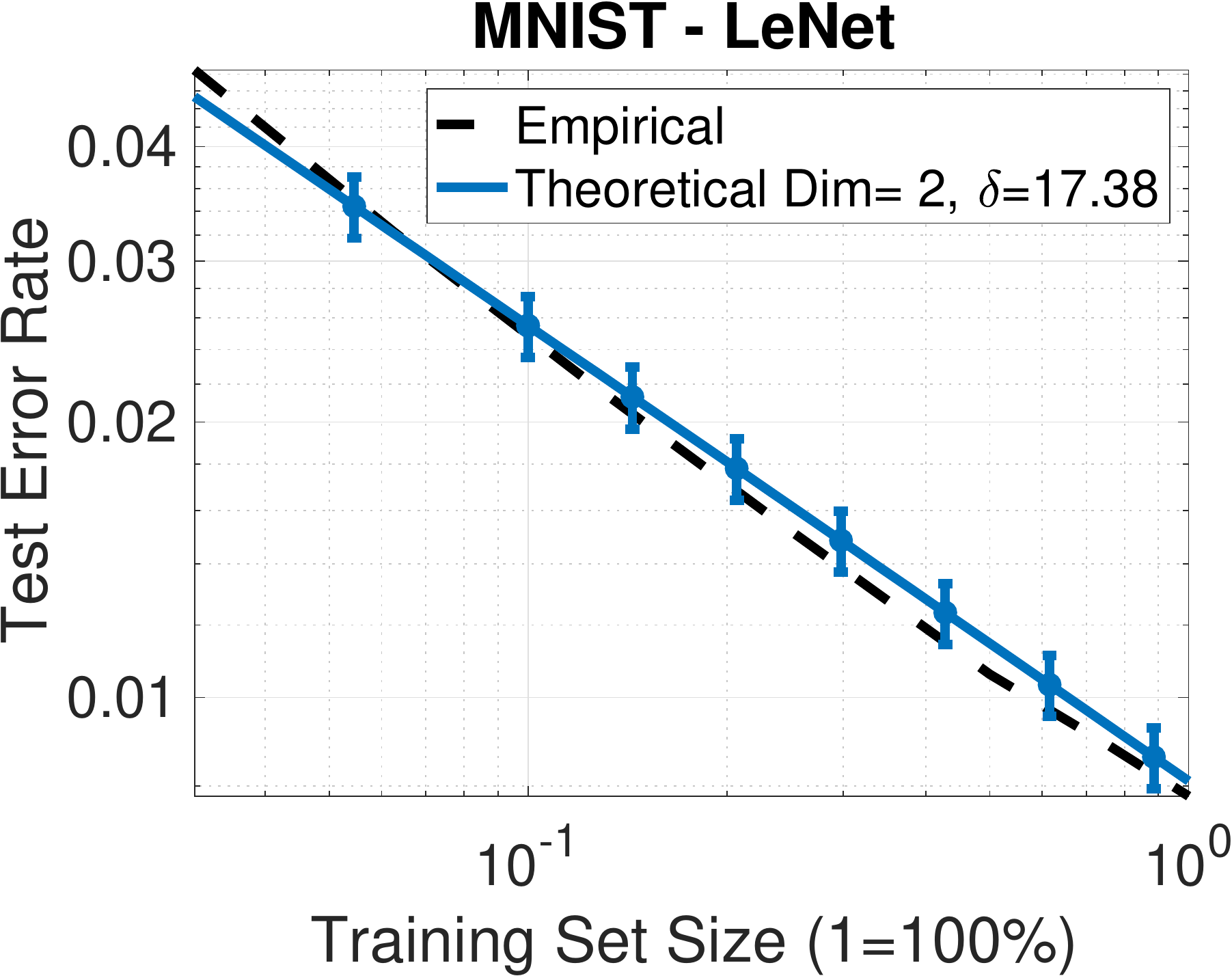}\\
    \vspace{5pt}
    \includegraphics[width=0.32\textwidth]{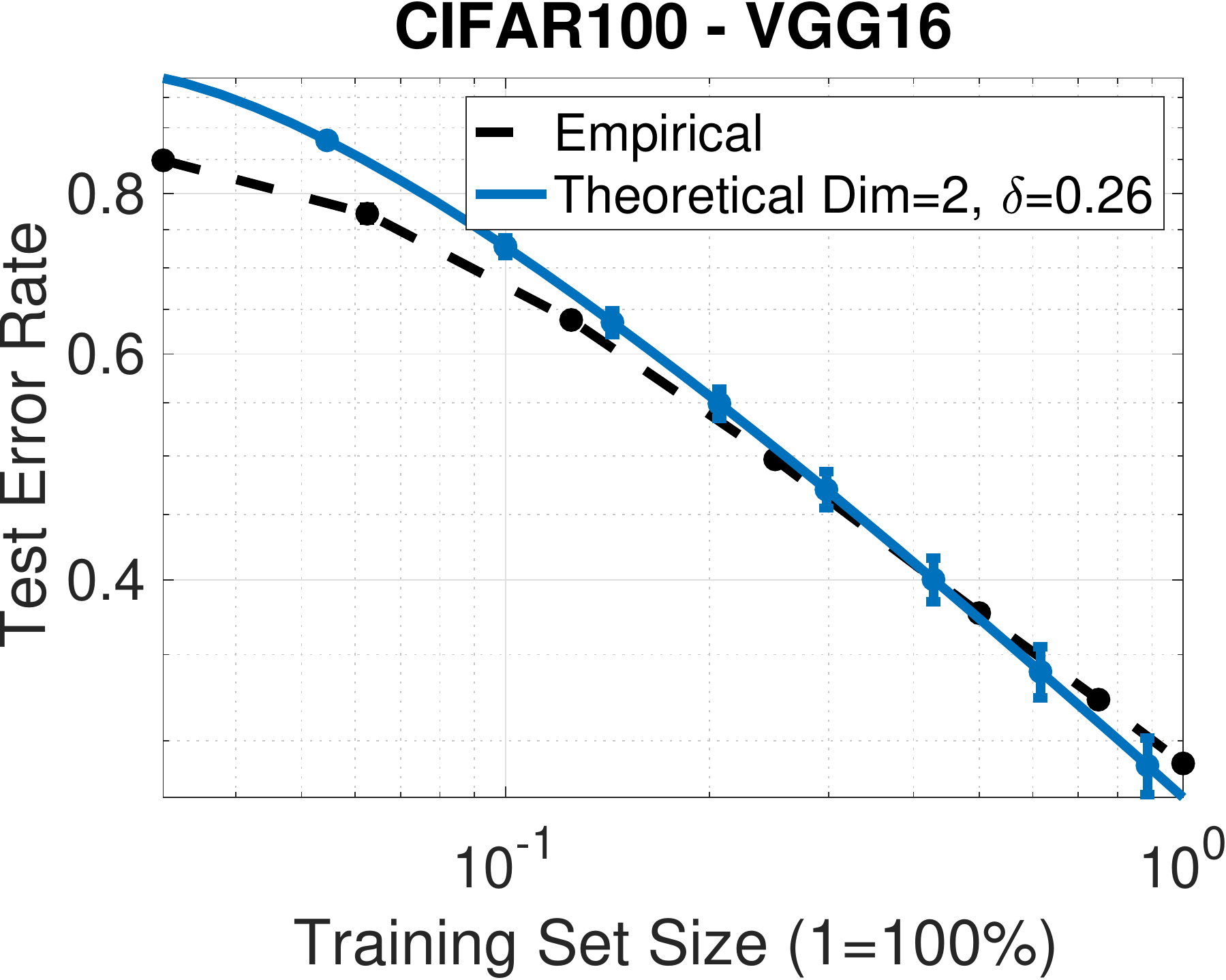}
    \includegraphics[width=0.32\textwidth]{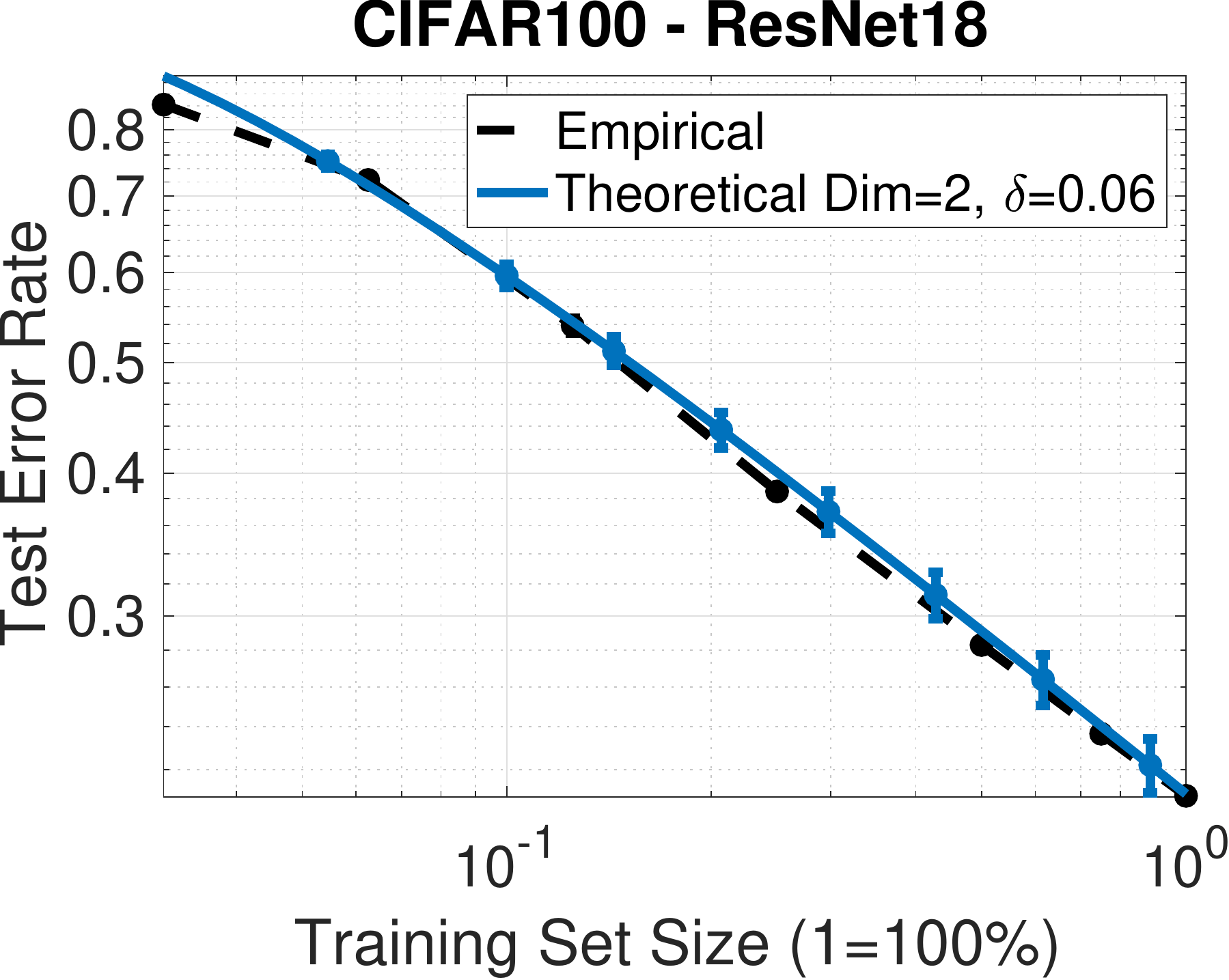}
    \includegraphics[width=0.32\textwidth]{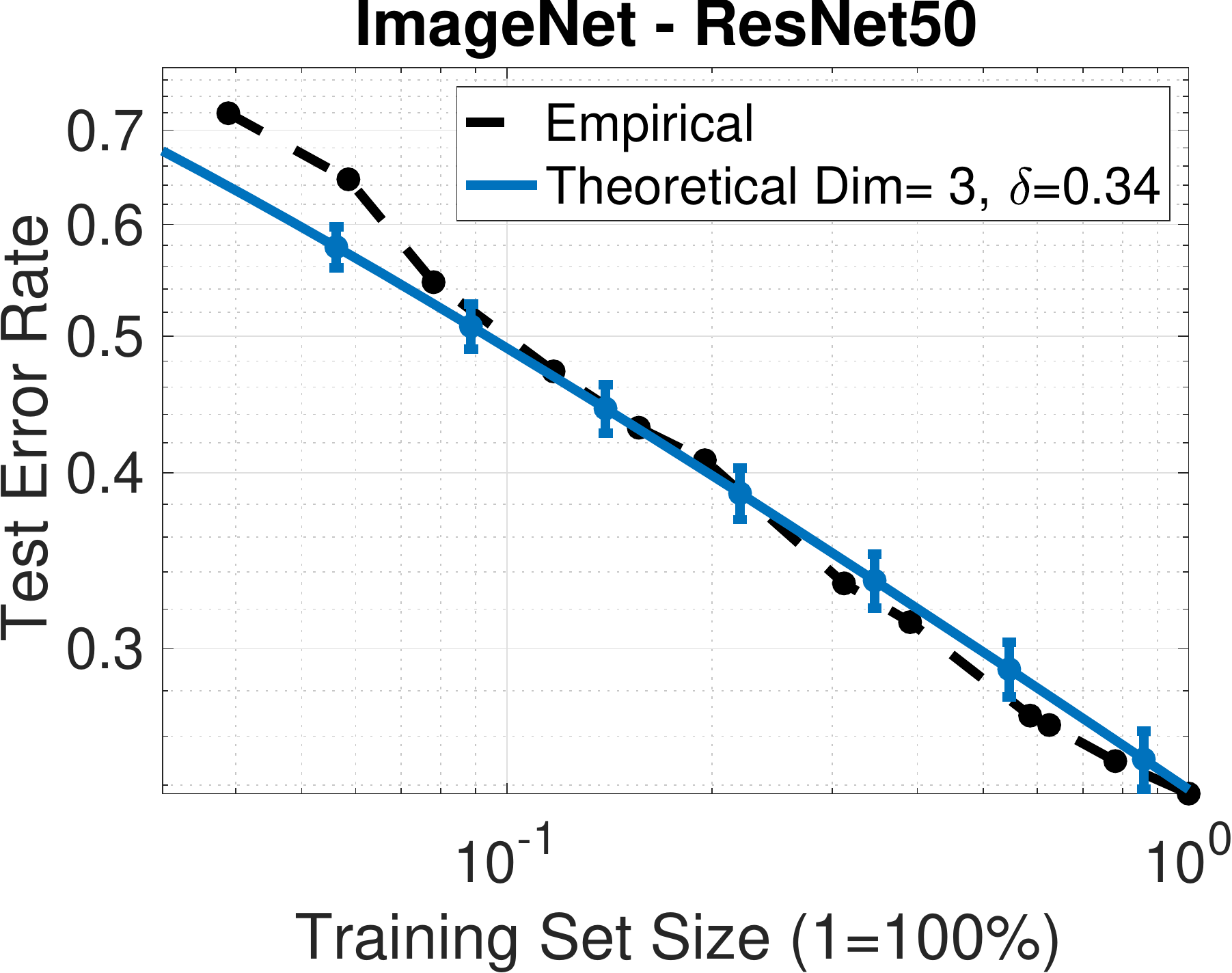}
    \vspace{5pt}
    \caption{Theoretical and empirical learning curves for classification experiments \textbf{Top Row:} CIFAR-10 trained on (\textbf{left}) VGG16, (\textbf{center}) ResNet-18 and (\textbf{right}) MNIST trained on LeNet.  \textbf{Bottom Row:} CIFAR-100 trained on \textbf{(left)}VGG16 and \textbf{(center)} ResNet-18, and \textbf{(right)} ImageNet trained on ResNet-50 .}
    \label{fig:class_results}
    \vspace{-10pt}
\end{figure*}

\begin{table}[h]
    \centering
    \begin{tabular}{|p{1.5cm}|M{1cm}|M{1cm}|M{1cm}|}
        \hline
        \multirow{2}{1.5cm}{\# filters in Conv1} & \multicolumn{3}{c|}{\# filters in Conv2} \\
        \cline{2-4}
        & 4 & 8 & 16  \\
        \hline
        2 & $3$ & $3$ & $3$ \\
        \hline
        4 & $2$ & $2$ & $2$ \\
        \hline
        6 & $2$ & $2$ & $2$ \\
        \hline
    \end{tabular}
    \caption{$d$ values for networks of varying capacity (i.e. varying number of filters in the first (Conv1) and second (Conv2) convolutional layer of the LeNet model.}
    \label{tab:mnist_lenet}
\end{table}
\begin{table}[h]
    \centering
    \begin{tabular}{|c||c|c|c|c|c|c|}
        \hline
        Width & 10 & 20 & 50 & 100 & 200 & 300\\
        \hline
        \hline
        d & 5 & 4 & 2 & 2 & 2 & 2\\
        \hline
    \end{tabular}
    \caption{$d$ values for networks of varying capacity (MLP with single hidden layer and varying width).}
    \label{tab:mnist_fcnet}
    \vspace{-15pt}
\end{table}
In Figure~\ref{fig:bottleneck} and Table~\ref{tab:sup_bottleneck} in the Supplement we show the experiment capturing the selection of the effective dimensionality involving the injection of the bottleneck to the network (it was described in the Section~\ref{sec:theoretical_framework}) for MNIST, CIFAR10, CIFAR100, and ImageNet data sets. Effective dimensionality $d$ is chosen as the size of the bottleneck for which we start observing saturation. We empirically found (see Figure \ref{fig:supp_class_results}) that this choice of $d$ allows to accurately estimate the learning curve, even when the accuracies of bottleneck models do not reach the accuracies of the original models as is the case for CIFAR100 and ImageNet data sets. Note that the accuracy of the model with the bottleneck saturates at $d=2$ for MNIST and CIFAR10, $d=2/d=3$ for CIFAR100 data set, and $d=3/d=4$ for ImageNet data set. 

Furthermore, we also extracted feature vectors of different dimensions from the bottleneck model and performed nearest neighbor classification on the low-dimensional features. We found that the performance of the nearest neighbor saturates for the same values of $d$ as described above. These results are highlighted in Figure~\ref{fig:supp_knn} and Table~\ref{tab:supp_knn} in the Supplement. This ensures us that the dimensionality found using the bottleneck indeed captures enough variety in the data to perform accurate prediction. Finally, for the Udacity, the effective dimensionality we found was equal to $1$.
Apart from training data set, the effective dimensionality of the feature space indirectly depends on the capacity of the neural network which in turn depends on network design. We verify this claim by training multiple LeNet and MLP models with varying capacity on MNIST data set and computing the effective dimensionality for each of the model. The LeNet model consists of two convolution layers with $6$ and $16$ filters respectively. We control the capacity of the network by decreasing the number of filters in each convolutional layer. For MLP, we use single hidden layer and vary its width. As the capacity of the network decreases we observe an increase in the effective dimensionality. The results are highlighted in Table~\ref{tab:mnist_lenet} and~\ref{tab:mnist_fcnet}.

\subsubsection{Learning curves}
\begin{figure}[h]
    \centering
    \includegraphics[width=0.35\textwidth]{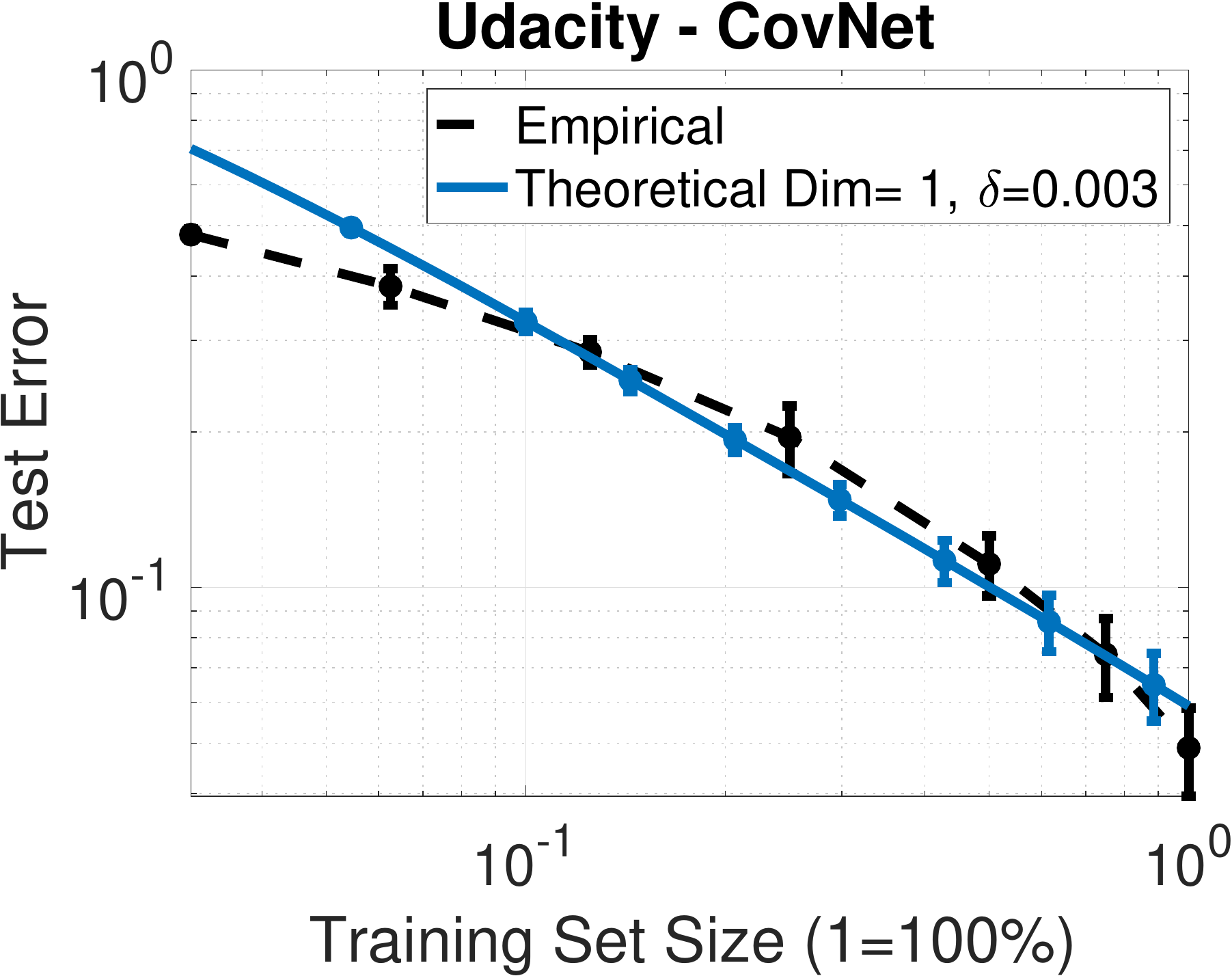}
    \caption{Theoretical and empirical learning curves for Udacity data set.}
    \label{fig:reg_results}
    \vspace{-15pt}
\end{figure}

\begin{table*}[!ht]
    \centering
    \begin{tabular}{|c|c|c|c|c|c|c|c|}
        \hline
        &&\multicolumn{2}{c|}{$f_{train}$} & \multicolumn{2}{c|}{$f_{test}$} && \\
        \cline{3-6}
        Data set & Model &  $\mu$ & diag($\Sigma$) & $\mu$ & diag($\Sigma$)& $d$ & $\delta$ \\ \hline
         MNIST & LeNet & \bm{0.020}{0.012} & \bm{377.855}{264.029} & \bm{-0.061}{-0.036} & \bm{384.357}{270.399} &  2 & 17.375  \\[2ex] \hline
        \multirow{2}*[-0.9em]{CIFAR-10} & VGG-16 & \bm{-0.004}{0.024} & \bm{98.192}{80.676} & \bm{0.009}{-0.060} & \bm{87.691}{76.163} & 2 & 0.890 \\[2ex] \cline{2-8}
        &  ResNet-18 & \bm{-0.003}{-0.001} & \bm{3.717}{2.789} & \bm{0.007}{0.003} & \bm{3.726}{2.662}& 2 & 0.262 \\[2ex] \hline
        \multirow{2}*[-0.9em]{CIFAR-100} & VGG-16 & \bm{0.023}{0.0464} & \bm{123.572}{121.108} & \bm{-0.0584}{-0.116} & \bm{102.838}{97.932} & 2 & 0.265 \\[2ex] \cline{2-8}
        & ResNet-18 & \bm{-0.001}{-0.010} & \bm{3.648}{3.449} & \bm{0.003}{0.0260} & \bm{2.914}{2.811} & 2 & 0.056 \\[2ex] \hline
        ImageNet & ResNet-50& \bmm{-0.002}{-0.007}{-0.020} &\bmm{19.940}{14.308}{12.331} &\bmm{ -0.002}{-0.007}{-0.020} &\bmm{19.940}{14.308}{12.331} & 3 & 0.340 \\[3.5ex] \hline
        Udacity & CovNet & $-0.0111$ & $4.3415$ & $0.0265$ & $6.1000$ & $1$ & $0.003$ \\ \hline
    \end{tabular}
    \caption{The effective dimensionality $d$ and $\delta$ parameter for different data sets and model architectures. The train and test feature distributions are denoted as $f_{train}$ and $f_{test}$. $\mu$ denotes the mean of the distribution and $diag(\Sigma)$ denotes the diagonal elements of the co-variance matrix (off-diagonal elements are equal to $0$).}
    \label{tab:sup_all_results}
    \vspace{-10pt}
\end{table*}
The empirical learning curves were obtained by testing DNNs trained on increasingly larger data sets. Thus we sampled MNIST, CIFAR and Udacity data set to obtain training data sets of size equal to $3.125\%$, $6.25\%$, $12.5\%$, $25\%$, $50\%$, $75\%$ and $100\%$ of the entire data set. For ImageNet we obtained training data sets of size equal to $3.90\%$, $5.85\%$, $7.80\%$, $11.70\%$, $15.61\%$, $19.51\%$, $31.22\%$, $39.02\%$, $58.54\%$, $62.44\%$, $78.05\%$ and $100\%$ of the entire data set. In the obtained training data sets, all classes are equally well-represented (they are balanced). For classification problems, we plot the experimental learning curve by computing the test error rate, i.e number of samples incorrectly classified by the model (see Figure~\ref{fig:class_results} and Table~\ref{tab:sup_all_results}; theoretical curves for various settings of effective dimensionality are reported in Figure~\ref{fig:supp_class_results} in the Supplement). For regression task, we count the test sample as misclassified if the predicted steering command deviates from the label by more than $0.1$ (in the Udacity the steering command is typically in the range (-0.5, 0.5); see Figure \ref{fig:reg_results} for the results). 

The theoretical learning curves were obtained according to Equations~\ref{eq:exp_phi_test_1d} and ~\ref{eqn:Exp_Phi_test_hd}, where the integral were computed using Monte Carlo method. Note that our generalization error estimate is dependent on the feature train and test distributions. In order to obtain the features for the distribution estimation, we train the DNN on a subset of the training data, i.e. $50\%$. Next, we process this subset as well as the subset of the test data with this DNN. The obtained features are then projected via PCA to the effective dimensionality $d$. We assume single $d$-dimensional Gaussian distribution for both the train and test data, whose parameters (mean and covariance) we estimate via maximum likelihood approach (it has been previously observed that features space learned by DNNs exhibit a simple clustering structure~\cite{goldfeld2018estimating}). Finally, we treat $\delta$ as a hyperparameter of the error estimate. It was obtained under small data regime ($\leq50\%$ of training data) by minimizing the distance between theoretical and empirical curve. Therefore, after training network on small amount of data, which is computationally much faster than training on the entire corpus, we estimate $\delta$ and predict the behavior of the learning curve in large data regime. Table~\ref{tab:sup_all_results} and~\ref{tab:my_label} in the Supplement summarizes the choice of hyperparameters for different data sets and architectures. As can be seen in Table~\ref{tab:sup_all_results}, $\delta$ heavily depends on the considered combination of data set and architecture (difference is often in order of magnitudes).

\begin{figure}[H]
    \centering
    \includegraphics[width=0.35\textwidth]{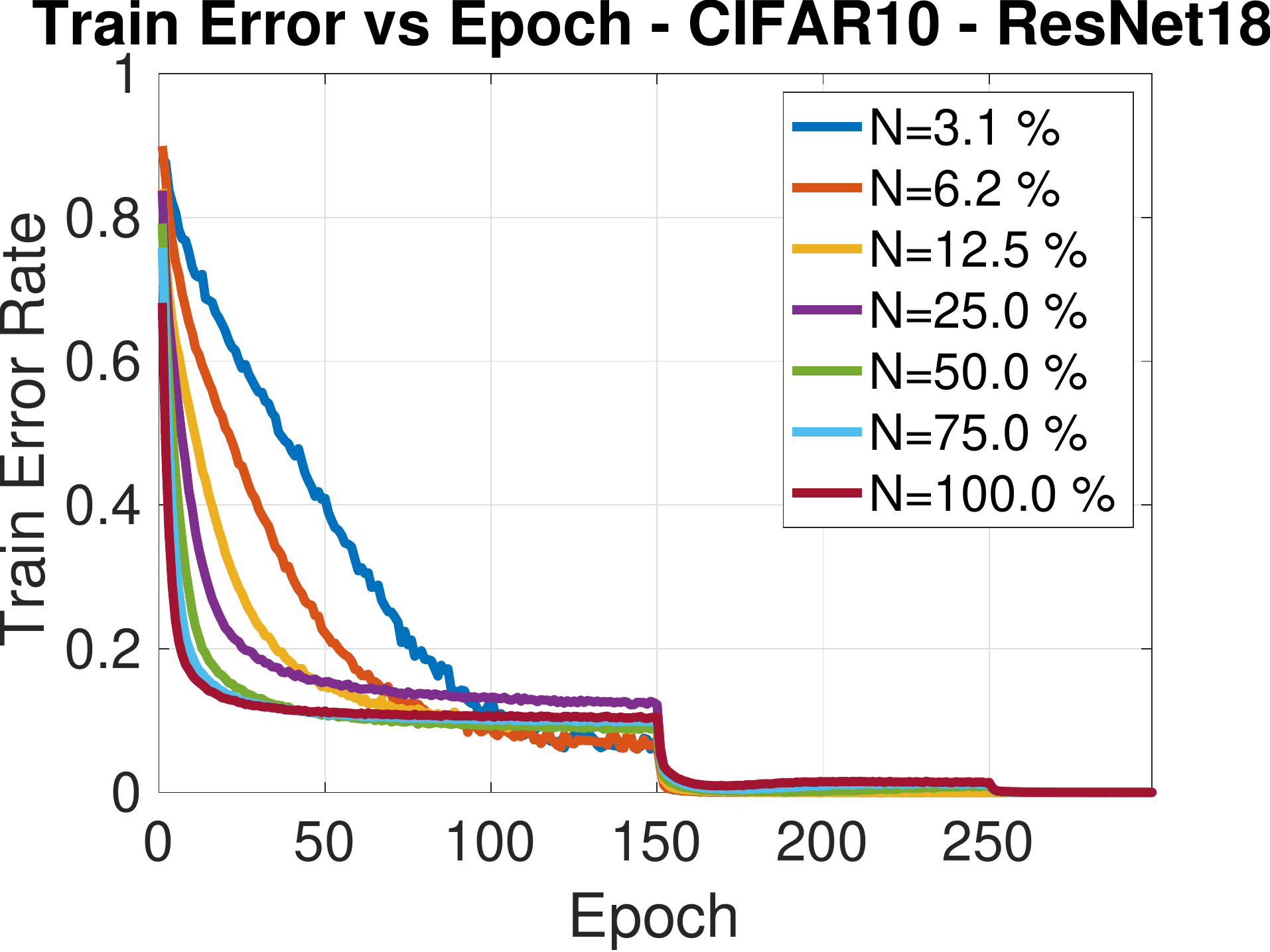}
    \includegraphics[width=0.35\textwidth]{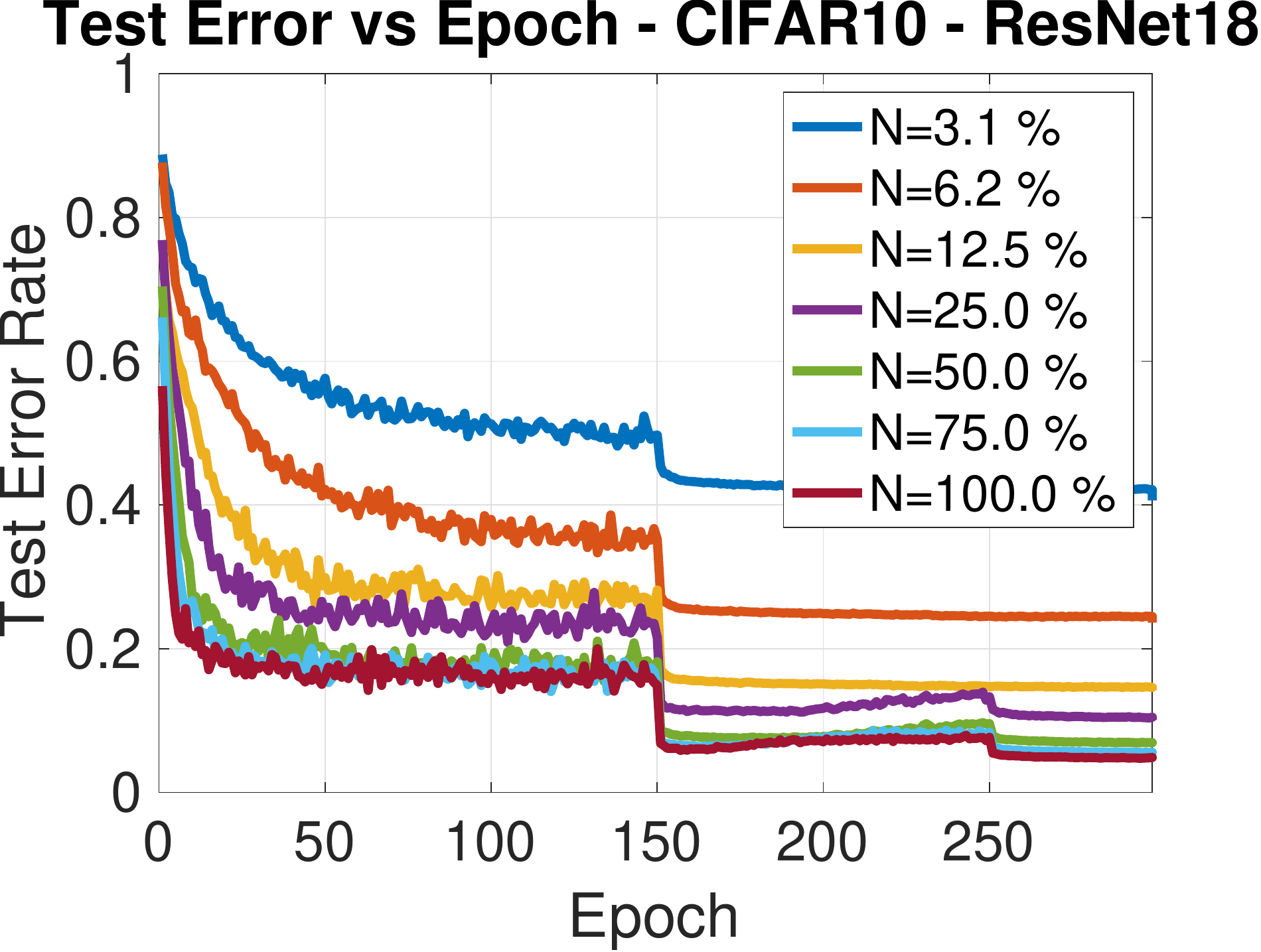}
    \includegraphics[width=0.35\textwidth]{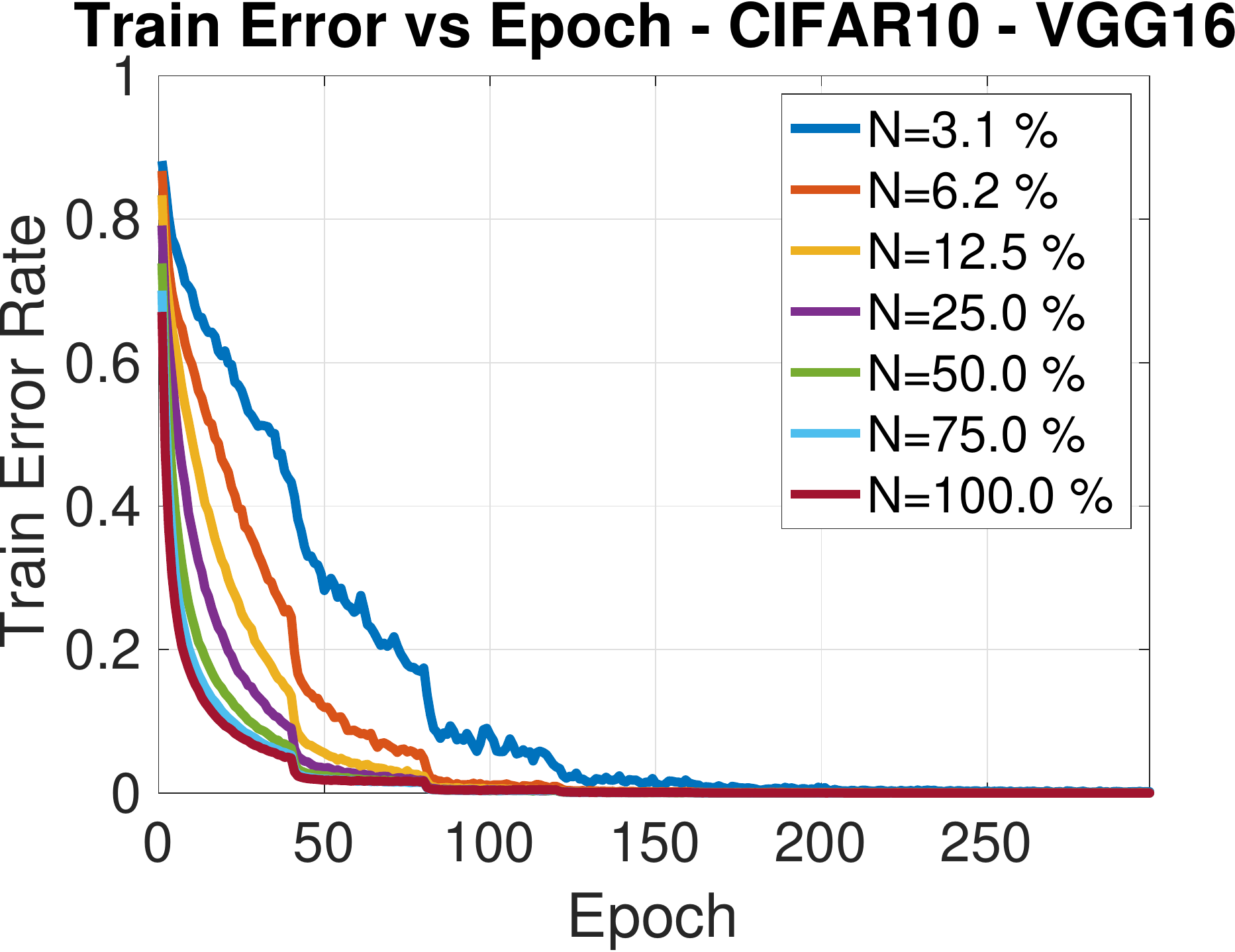}
    \includegraphics[width=0.35\textwidth]{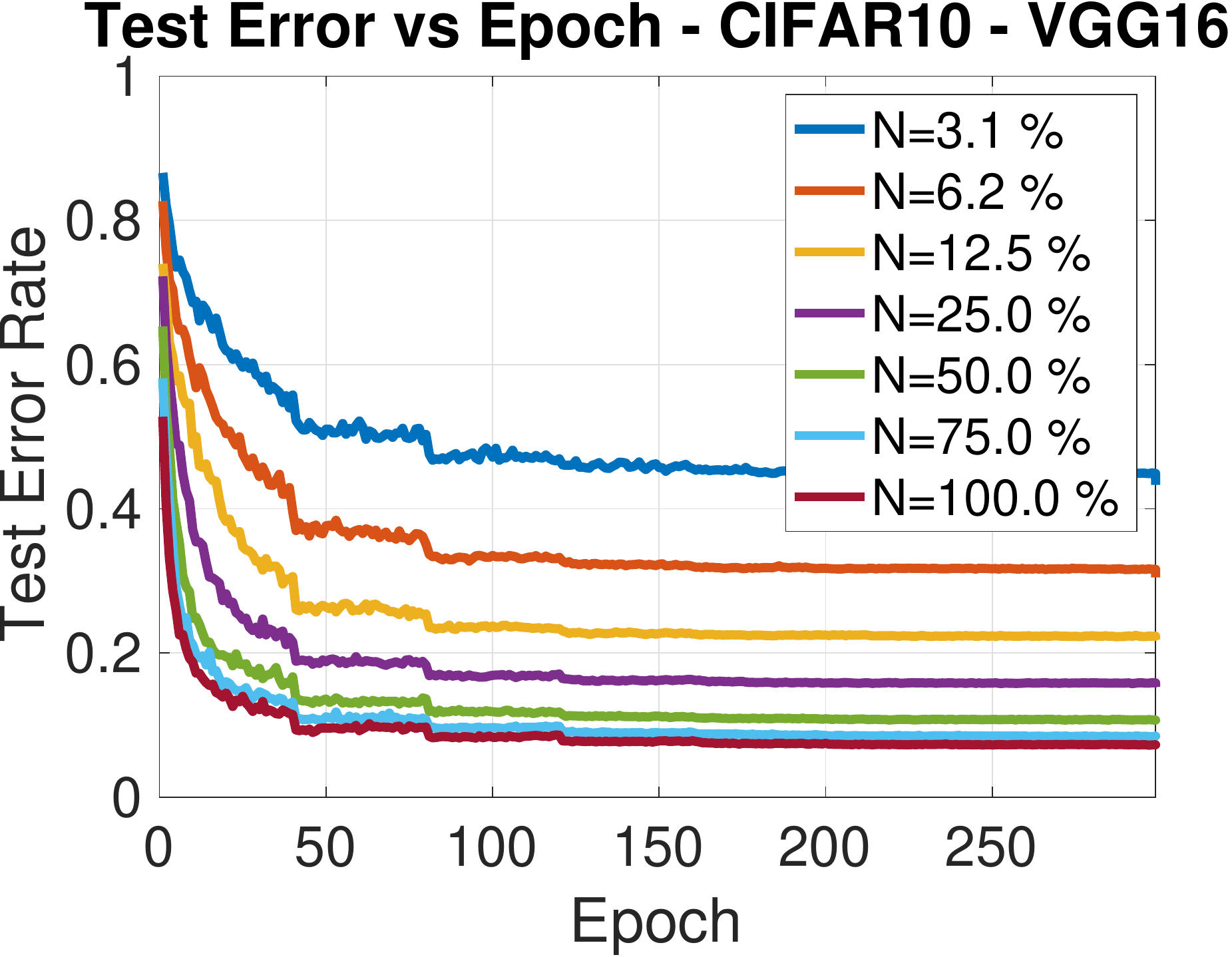}
    \caption{\textbf{(Left:)} Train and \textbf{(Right:)} test error vs number of epochs for \textbf{(Top:)} ResNet-18 \textbf{(Bottom:)} VGG16 models trained on increasingly larger subsets of CIFAR-10 data set.}
    \label{fig:cifar10}
\end{figure}

\begin{figure}[H]
    \centering
    \includegraphics[width=0.35\textwidth]{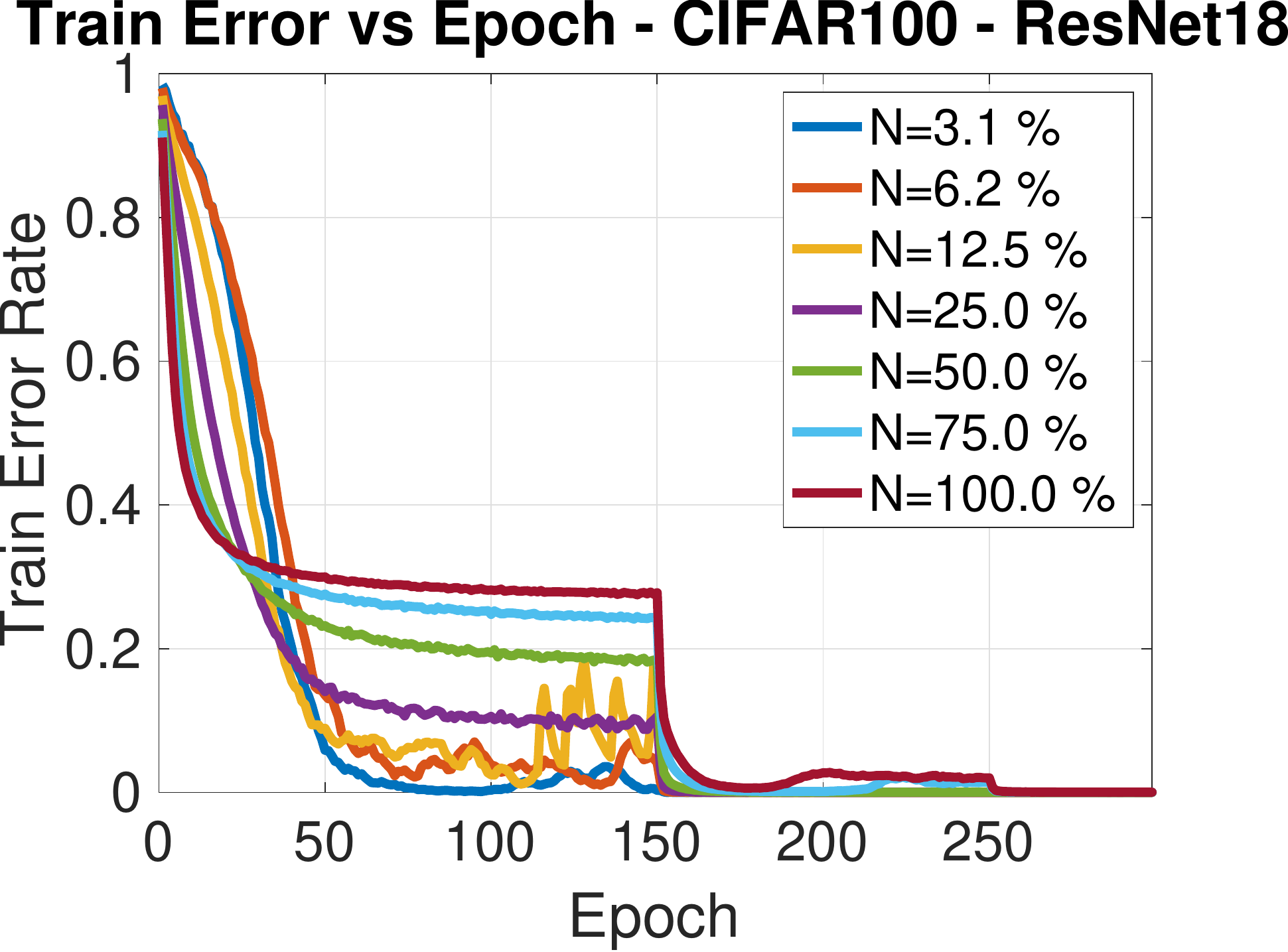}
    \includegraphics[width=0.35\textwidth]{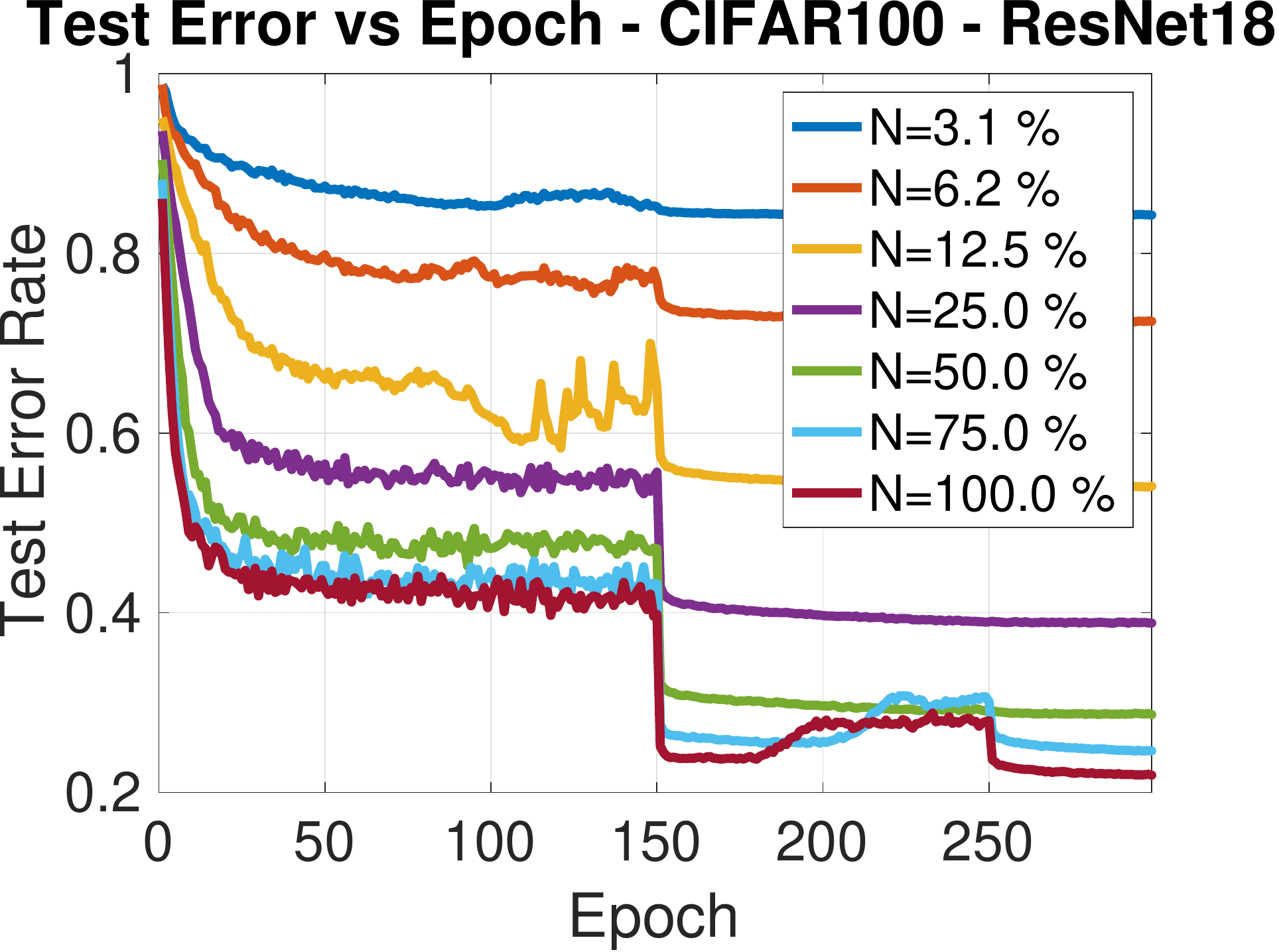}
    \includegraphics[width=0.35\textwidth]{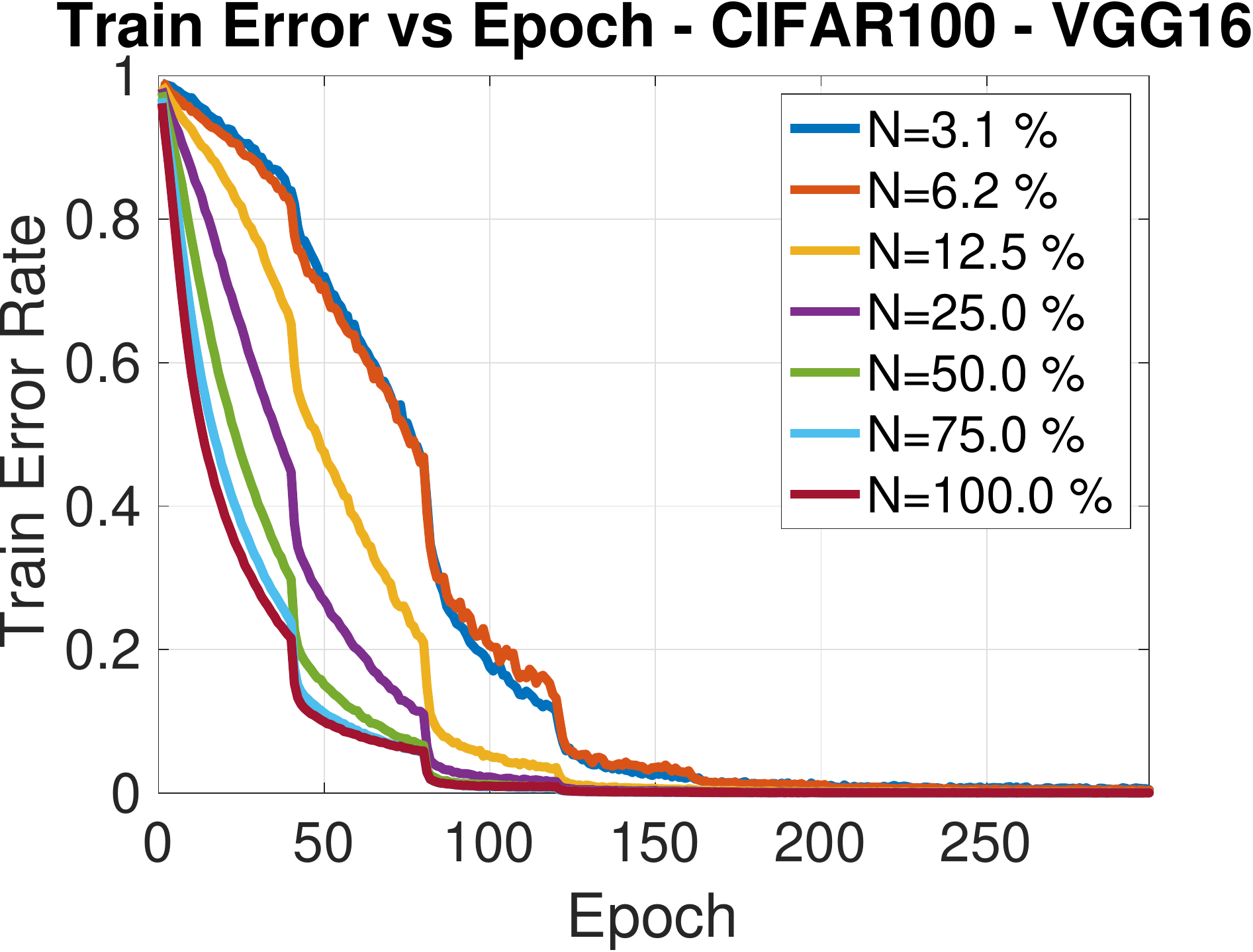}
    \includegraphics[width=0.35\textwidth]{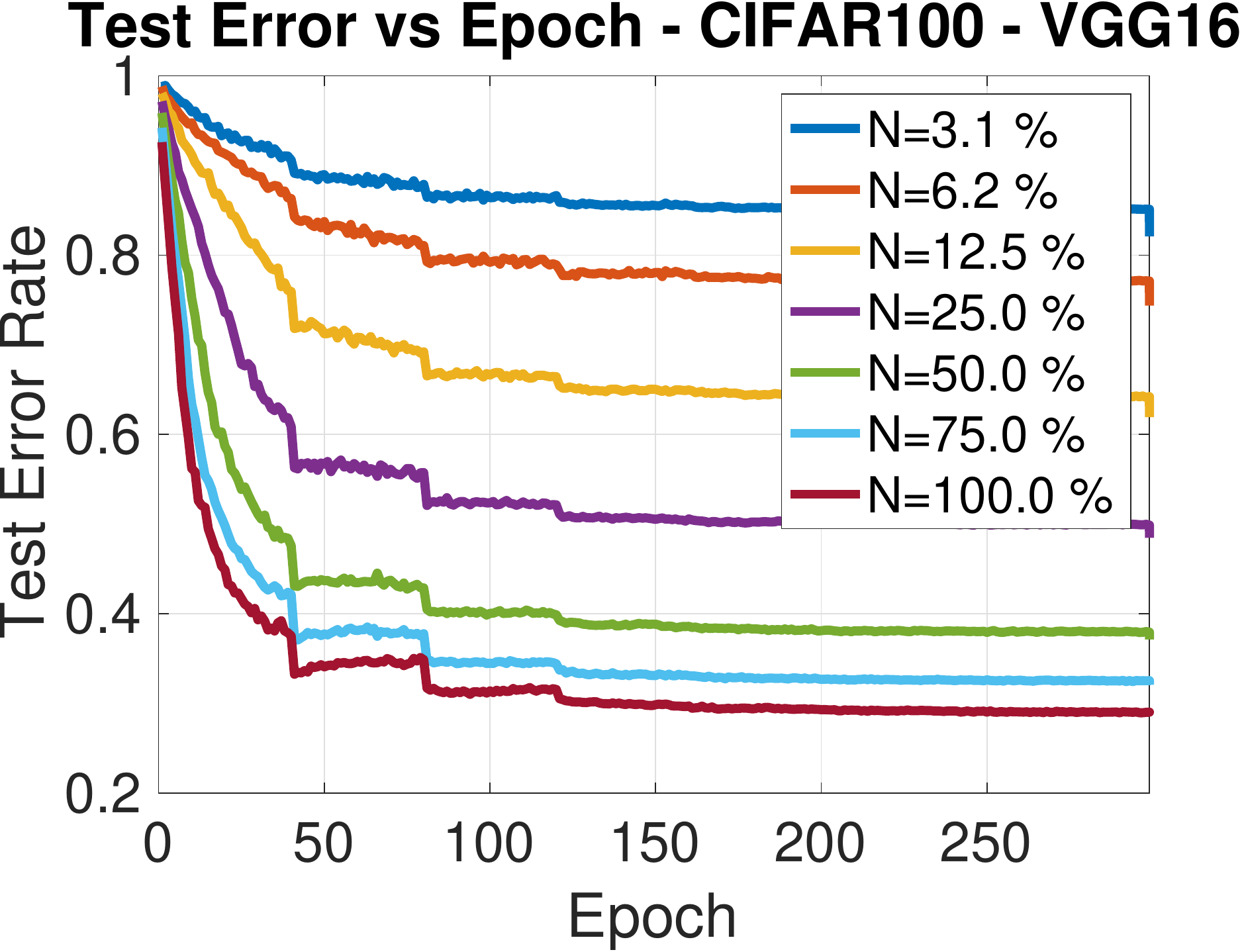}
    \caption{\textbf{(Left:)} Train and \textbf{(Right:)} test error vs number of epochs for \textbf{(Top:)} ResNet-18 \textbf{(Bottom:)} VGG16 models trained on increasingly larger subsets of CIFAR-100 data set.}
    \label{fig:cifar100}
\end{figure}

\begin{figure}[H]
    \centering
    \includegraphics[width=0.42\textwidth]{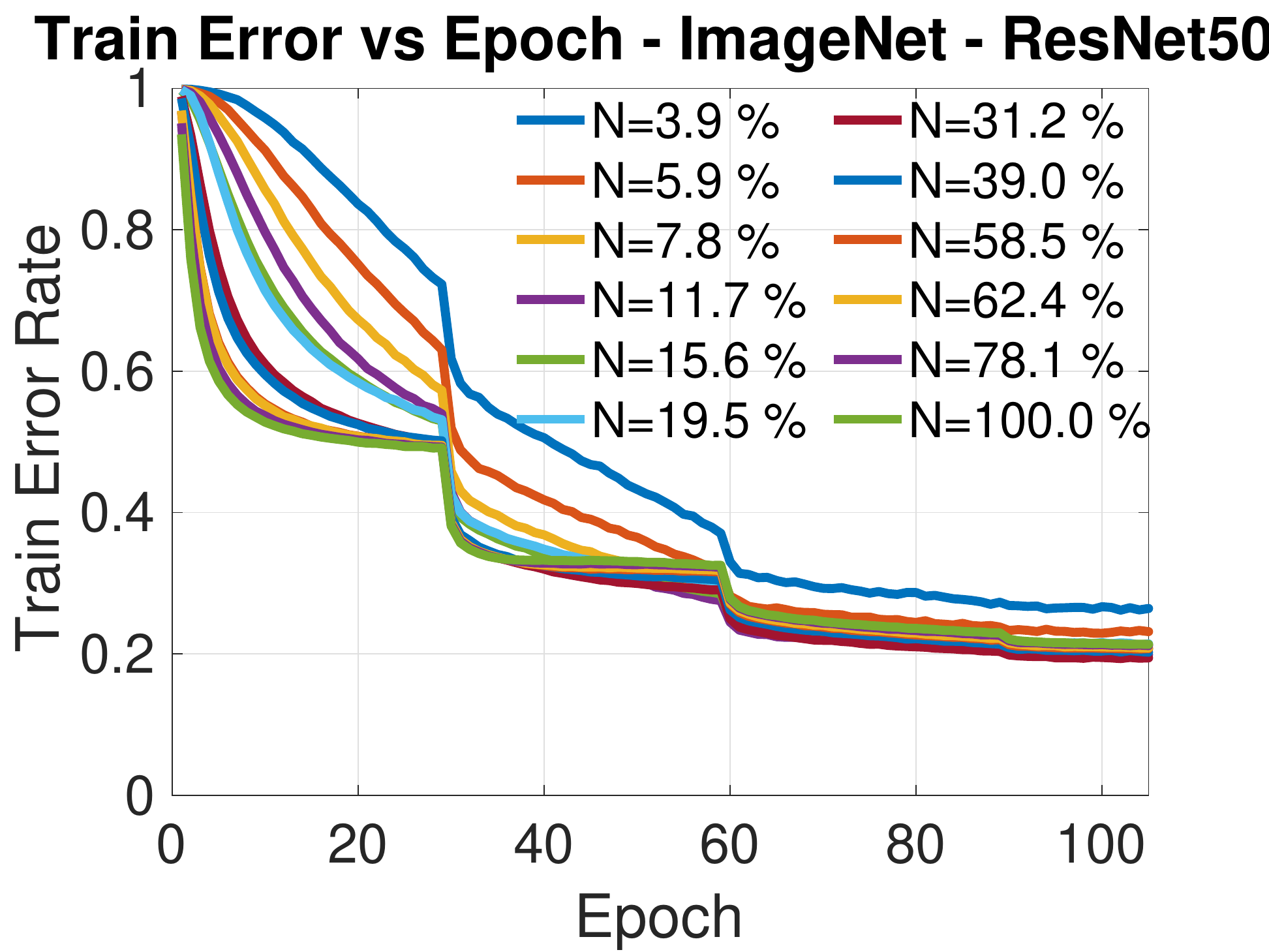}
    \includegraphics[width=0.42\textwidth]{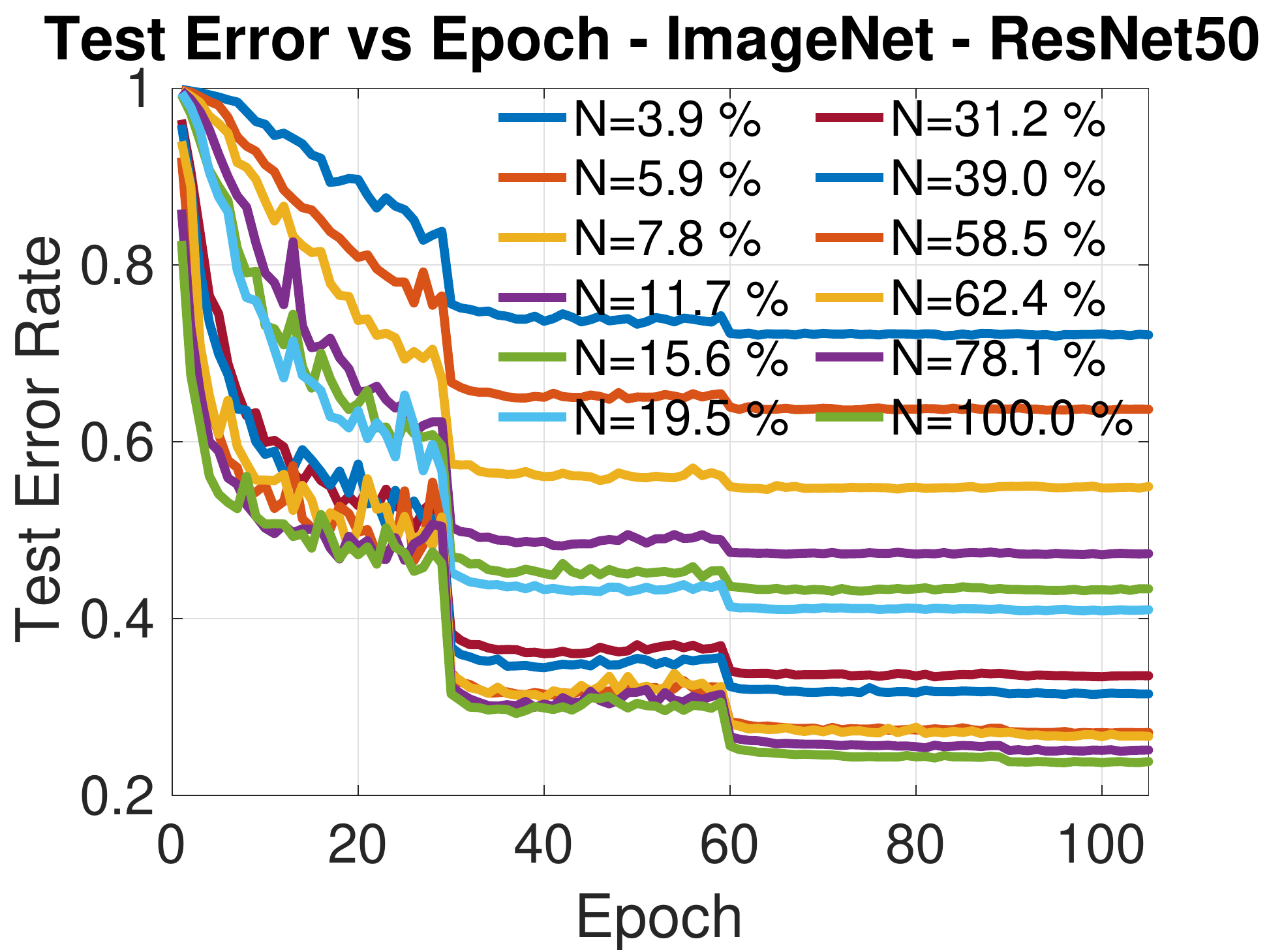}
    \caption{\textbf{(Left:)} Train and \textbf{(Right:)} test error vs number of epochs for ResNet-50 model trained on increasingly larger subsets of ImageNet data set.}
    \label{fig:imagenet}
\end{figure}

\begin{figure}[H]
    \centering
    \includegraphics[width=0.42\textwidth]{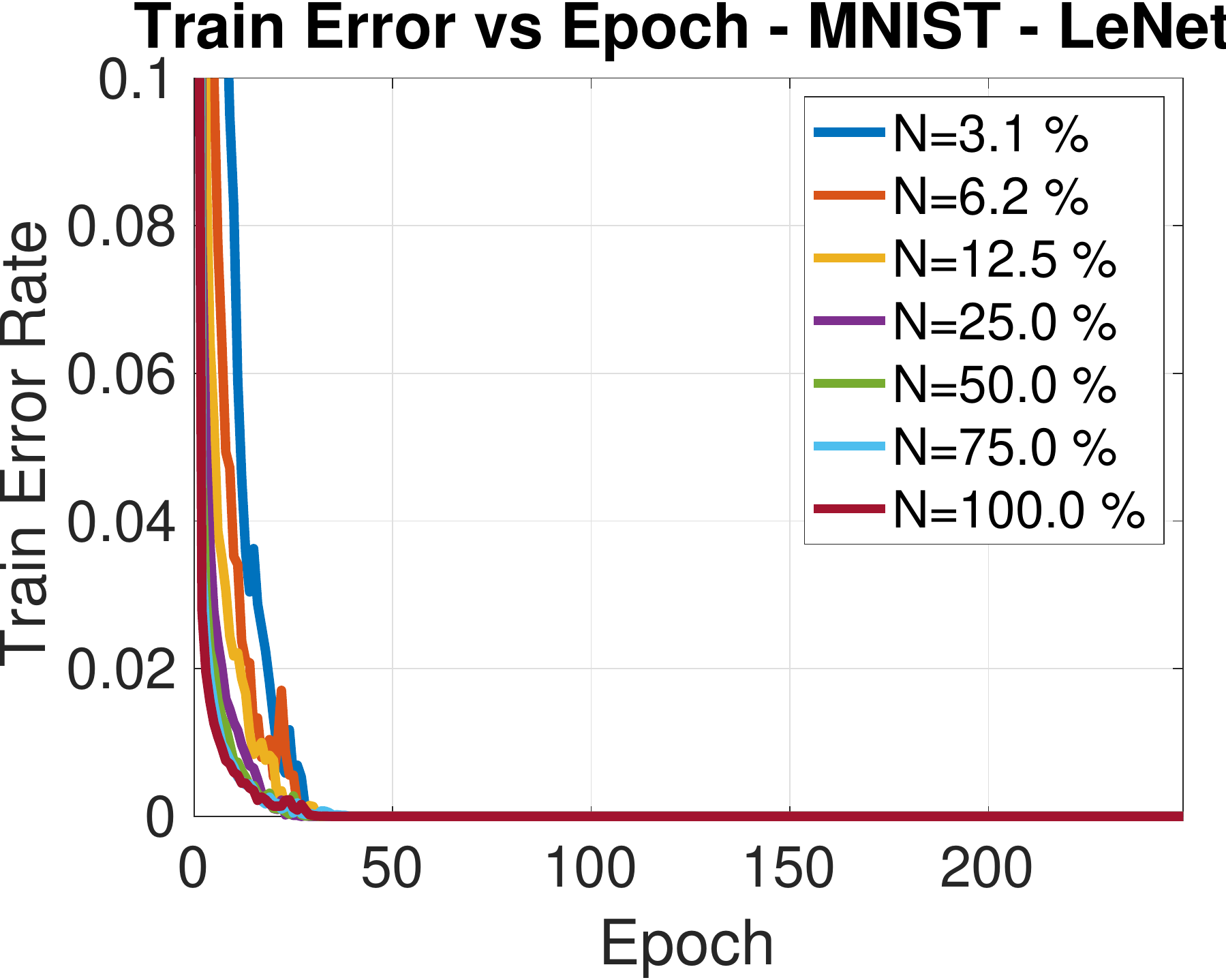}
    \includegraphics[width=0.42\textwidth]{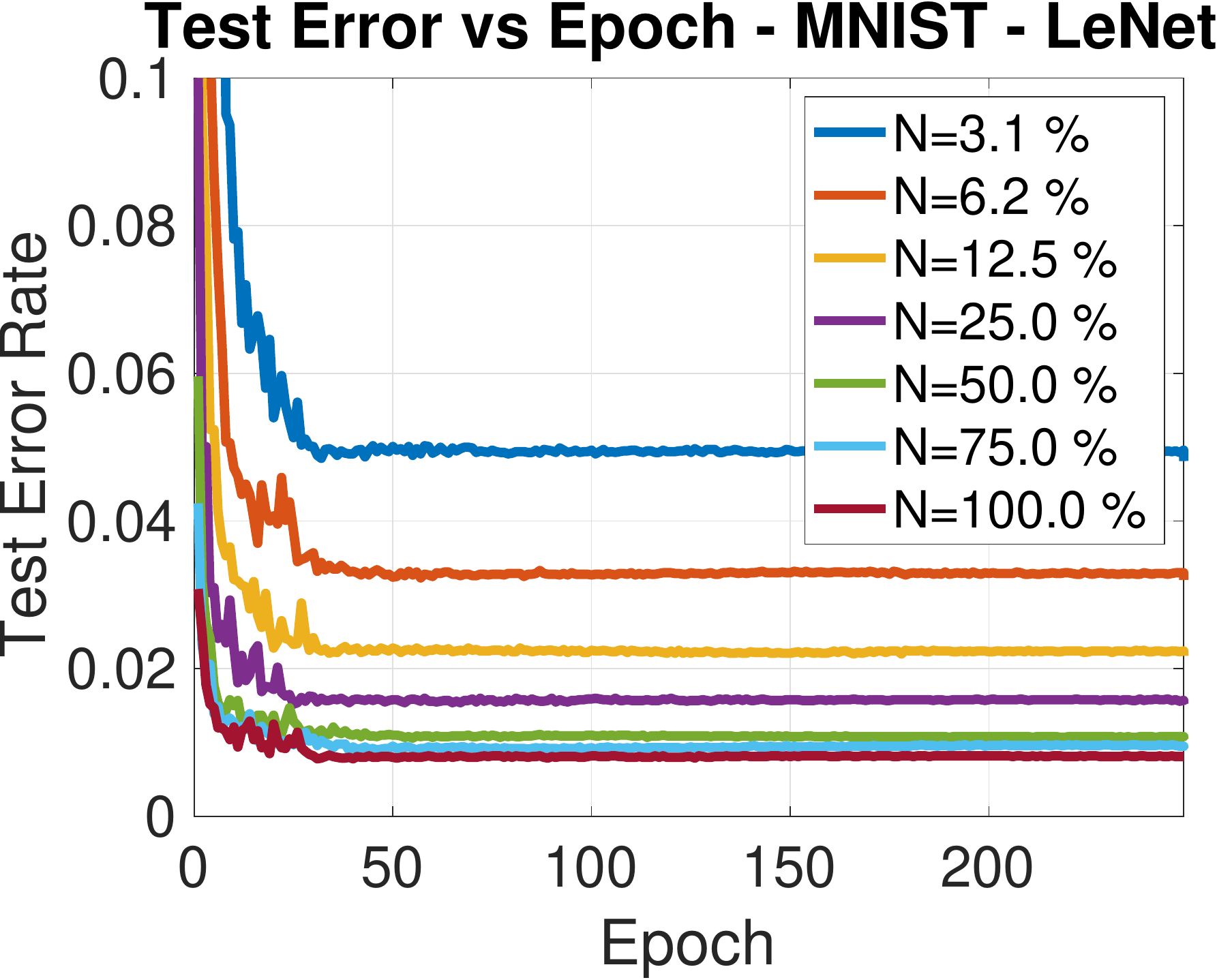}
    \caption{\textbf{(Left:)} Train and \textbf{(Right:)} test error vs number of epochs for LeNet models trained on increasingly larger subsets of MNIST data set.}
    \label{fig:mnist}
\end{figure}

\begin{figure}[H]
    \centering
    \includegraphics[width=0.45\textwidth]{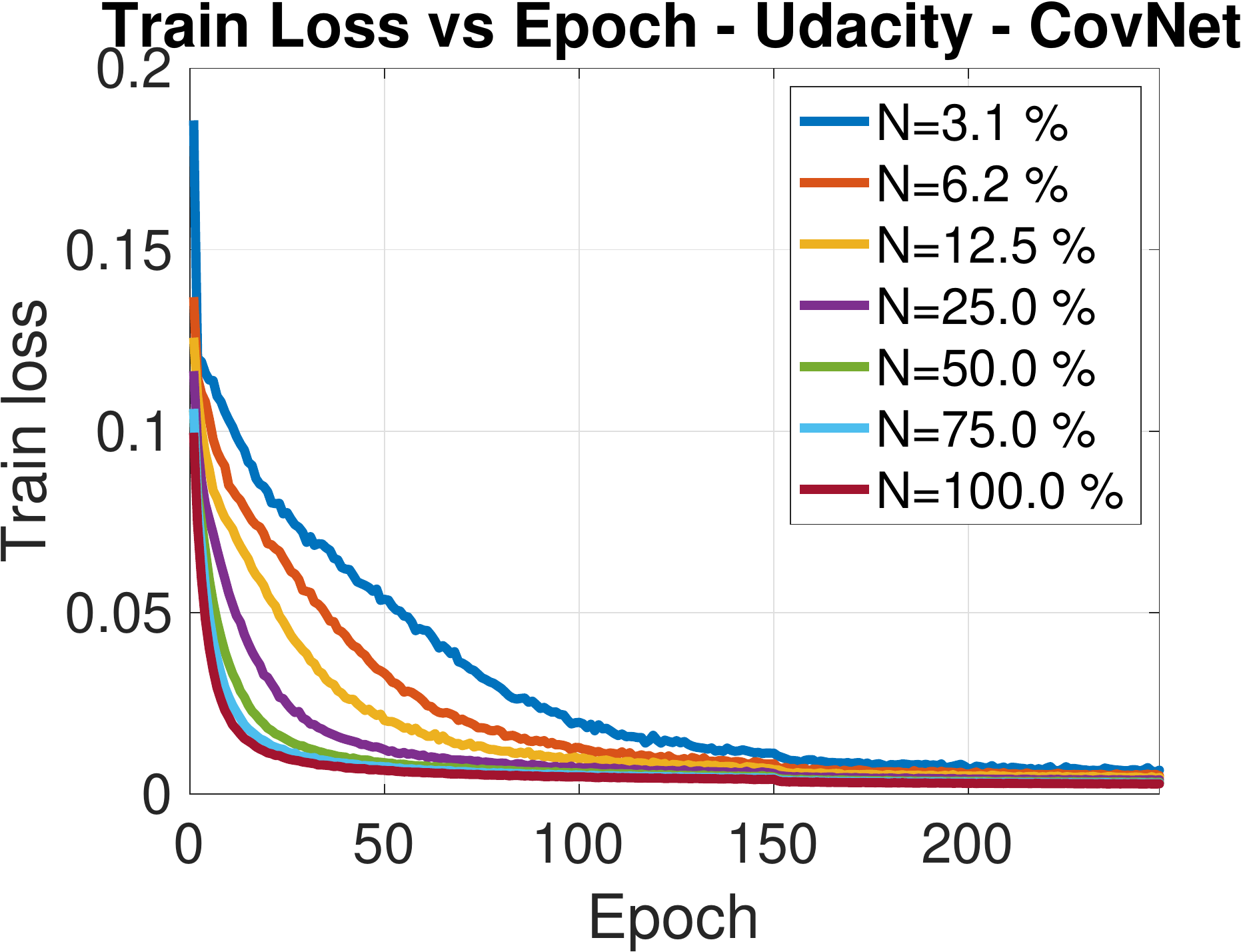}
    \includegraphics[width=0.45\textwidth]{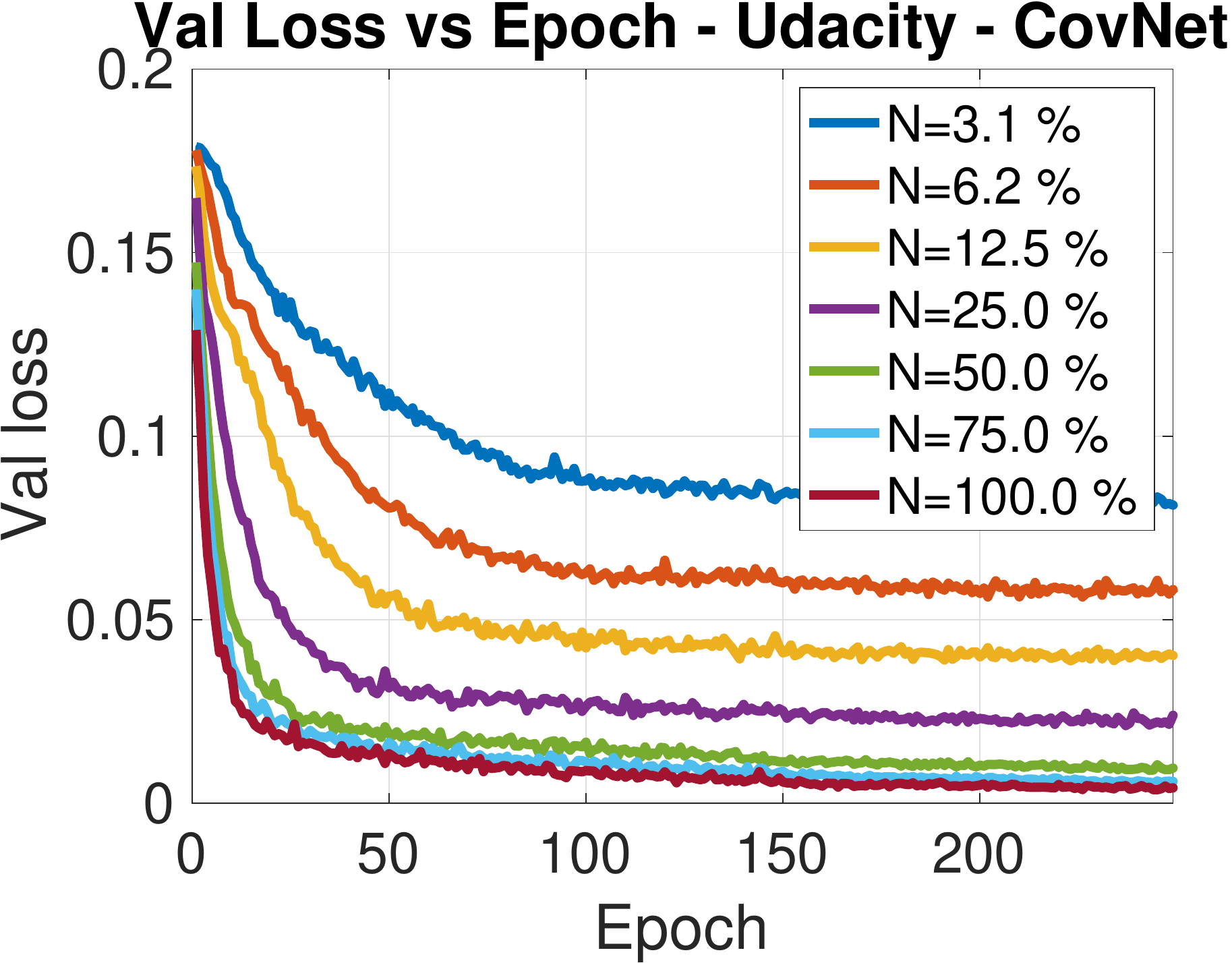}
    \caption{\textbf{(Left:)} Train and \textbf{(Right:)} test error vs number of epochs for CNN models trained on increasingly larger subsets of Udacity data set.}
    \label{fig:udacity_curves}
\end{figure}

Figure ~\ref{fig:class_results}, Figure~\ref{fig:reg_results} and Table~\ref{tab:sup_all_results} report the results confronting the theoretical and empirical learning curves. Note that among all our data sets, only ImageNet does not satisfy the assumption of zero training error (see Figures~\ref{fig:mnist},~\ref{fig:cifar10},~\ref{fig:cifar100},~\ref{fig:imagenet}; for Udacity data set the training error is close to zero as can be seen in Figure~\ref{fig:udacity_curves}), nevertheless even for this data set we could well-model the behavior of the learning curve using our theoretical framework. According to~\cite{DBLP:journals/corr/abs-1712-00409} the learning curve can be broken down into three regions: low data region, power law region, and the saturation region. In our experiments we observe first two regions. In low data regime we observe over-fitting. In this case we observe a mismatch between the theoretical and empirical curves (recall that our estimates of the generalization error become more accurate with increasing $N$). In the power law region, as we increase the amount of training data the performance of the network consistently improves. Our theoretical framework estimates the empirical learning curve in this region very well. 

\section{Conclusion}\label{sec:conclusion}
In this paper we address the problem of describing the behavior of the generalization error of DL models with the growing size of the training data. We attempt to reconcile the dichotomy between existing theoretical approaches, which rely on capacity measures that are potentially impossible to obtain for practical DNNs, and existing empirical approaches that model the behavior of the error by fitting it to a parameterized curve and lack any theoretical description. Our error estimates stem from a simple model of a DL machine that we propose and analyze. Our approach relies on modeling assumptions, which are however not unrealistic and gives rise to the estimates of the generalization error curves that closely resemble the ones empirically observed. We verify our approach on several learning tasks involving various realistic architectures and data sets.
\clearpage
{\small
\bibliographystyle{ieee_fullname}
\bibliography{egbib}
}
\clearpage \newpage

\onecolumn
\begin{center}
\textbf{\Large A Theoretical‐Empirical Approach to Estimating Sample Complexity of DNNs\\ (Supplementary Material)}
\end{center}

\section{Derivations for Equation~\ref{eq:exp_psi_1d}}

\begin{align*}
    &\mathbb{E}_u^{\langle x_{i}, x_{j} \rangle}[\psi(\hat{x})] = \int\displaylimits_{x_{i}}^{x_{j}}|\hat{x}-x(\hat{x})|u(\hat{x})d\hat{x} = \!\!\!\! \int\displaylimits_{x_{i}}^{\frac{x_{i} + x_{j}}{2}} \!\!\!\! (\hat{x}-x_{i})u(\hat{x})d\hat{x} \, + \!\!\!\! \int\displaylimits_{\frac{x_{i} + x_{j}}{2}}^{x_{j}}\!\!\!\!(x_{j} - \hat{x})u(\hat{x})d\hat{x} \\
    &=  \int\displaylimits_{x_{i}}^{x_{i} + \frac{\rho(\hat{x})}{2}}\!\!\!(\hat{x}-x_{i})u(\hat{x})d\hat{x} \, + \!\! \int\displaylimits_{x_{i} + \frac{\rho(\hat{x})}{2}}^{x_{i} + \rho(\hat{x})}\!\!(x_{i} + \rho(\hat{x})-\hat{x})u(\hat{x})d\hat{x} \\
    &= \! \frac{1}{\rho(\hat{x})}\Bigg(\int\displaylimits_{x_{i}}^{x_{i} + \frac{\rho(\hat{x})}{2}}\!\!\!(\hat{x}-x_{i})d\hat{x} \, +  \int\displaylimits_{x_{i} + \frac{\rho(\hat{x})}{2}}^{x_{i} + \rho(\hat{x})}\!\!\!(x_{i} + \rho(\hat{x})-\hat{x})d\hat{x} \Bigg) \\
    &= \frac{1}{\rho(\hat{x})}\Bigg(\Bigg[\frac{\hat{x}^2}{2}-x_{i}\hat{x}\vc{\Bigg]}{x_{i}}{x_{i} + \frac{\rho(\hat{x})}{2}} \!\!\!\!\!\!\!\! + \Bigg[x_{i}\hat{x} + \rho(\hat{x})\hat{x} - \frac{\hat{x}^2}{2}\!\!\vc{\Bigg]}{x_{i} + \frac{\rho(\hat{x})}{2}}{x_{i} + \rho(\hat{x})}\Bigg) \\
    &= \!\!\frac{1}{\rho(\hat{x})}\Bigg(\!\!\!-\frac{x_{i}\rho(\hat{x})}{2} \!+\! \frac{x_{i}\rho(\hat{x})}{2} \!+\! \frac{\rho(\hat{x})^2}{2} \!+\!  \frac{1}{2} \Big(x_{i} \!+\! \frac{\rho(\hat{x})}{2} \Big)^2 - \frac{x_{i}^2}{2} - \frac{(x_{i} + \rho(\hat{x}))^2}{2} \frac{1}{2}\left(x_{i} + \frac{\rho(\hat{x})}{2}\right)^2\Bigg) \\
    &= \frac{1}{\rho(\hat{x})}\Bigg(\frac{\rho(\hat{x})^2}{2} \!+\! \frac{x_{i}\rho(\hat{x})}{2} \!+\! \frac{\rho(\hat{x})^2}{8} - \frac{x_{i}^2}{2} - x_{i}\rho(\hat{x}) - \frac{\rho(\hat{x})^2}{2} \!+\! \frac{x_{i}^2}{2} \!+\! \frac{x_{i}\rho(\hat{x})}{2} \!+\! \frac{\rho(\hat{x})^2}{8}\Bigg) \\ 
    &= \frac{\rho(\hat{x})}{4} \qedhere
\end{align*}

\section{Real Data Experiments}
\subsection{Finding effective dimensionality}
\begin{table}[H]
    \centering
    \begin{tabular}{|c|c|c|c|c|c|c|c|c|}
        \hline
        \multirow{2}{*}{Data set} & \multirow{2}{*}{Architecture} & \multirow{2}{*}{Baseline}  & \multicolumn{6}{c|}{Bottleneck width ($d'$)}  \\
        \cline{4-9}
        & & & 1 & 2 & 3 & 4 & 5 & 6 \\ \hline
        MNIST & LeNet & 0.992 & 0.407 & 0.971 & 0.986 & 0.988 & 0.989 & 0.989 \\ \hline
        \multirow{2}{*}{CIFAR-10}  & ResNet18 & 0.955 & 0.842 & 0.945 & 0.950 & 0.954 & 0.948 & 0.953 \\ \cline{2-9}
                                  & VGG16 & 0.930 & 0.886 & 0.926 & 0.926 & 0.928 & 0.925 & 0.927 \\ \hline
        \multirow{2}{*}{CIFAR-100} & ResNet18 & 0.791 & 0.172 & 0.590 & 0.699 & 0.729 & 0.739  & 0.735 \\ \cline{2-9}
                                  & VGG16 & 0.712 & 0.161 & 0.610 & 0.672 & 0.683 & 0.689 & 0.695 \\ \hline
        ImageNet & ResNet 50 & 0.763 & 0.018 & 0.272 & 0.626 & 0.674 & 0.686 & 0.697 \\ \hline
        Udacity & CovNet & 0.950 & 0.9786 & 0.983 & 0.992 & 0.991 & 0.989 & 0.992 \\ \hline
    \end{tabular}
    \caption{Test accuracy captured for baseline model (model without bottleneck trained on full training data set) and models with different widths ($d' = \{1,2,3,4,5,6\}$) of the bottleneck.}
    \label{tab:sup_bottleneck}
\end{table}

We perform nearest neighbor classification on the low-dimensional features extracted from the bottleneck model described in Section~\ref{sec:d_eff}. The effective dimensionality we find using nearest neighbor framework for different data sets is as follows: $d=2$ for CIFAR-10 and MNIST, $d=2/3$ for CIFAR-100 and $d=3/4$ for ImageNet, and it is consistent with the effective dimensionality found in the bottleneck experiments. 

\begin{table*}[ht]
    \centering
    \begin{tabular}{|c|c|c|c|c|c|c|c|c|}
        \hline
        \multirow{2}{*}{Data set} & \multirow{2}{*}{Model} & \multirow{2}{*}{Baseline}  & \multicolumn{6}{c|}{Feature vector size}  \\
        \cline{4-9}
        & & & 1 & 2 & 3 & 4 & 5 & 6\\ \hline
        MNIST & LeNet & 0.992 & 0.3922 & 0.970 & 0.986 & 0.987 & 0.988 & 0.989 \\ \hline
        \multirow{2}{*}{CIFAR-10}  & ResNet18 & 0.955 & 0.838 & 0.944 & 0.950 & 0.952 & 0.951 & 0.952 \\ \cline{2-9}
                                   & VGG16 & 0.930 & 0.798 & 0.923 & 0.924 & 0.921 & 0.923 & 0.927 \\ \hline
        \multirow{2}{*}{CIFAR-100} & ResNet18 & 0.791 & 0.205 & 0.579 & 0.664 & 0.688 & 0.709 & 0.718  \\ \cline{2-9}
                                   & VGG16 & 0.712 & 0.169 & 0.575 & 0.643 & 0.645 & 0.668 & 0.677  \\ \hline
        ImageNet & ResNet-50 & 0.763 & 0.0153 & 0.311 & 0.585   & 0.626 & 0.642   & 0.649 \\ \hline
    \end{tabular}
    \caption{Nearest neighbor classification accuracy for different sizes of feature vectors ($d'=1,2,3,4,5,6$). Baseline model is the original DNN without bottleneck trained on full training data set.}
    \label{tab:supp_knn}
\end{table*}

\begin{figure}[H]
    \centering
    \includegraphics[width=0.32\textwidth]{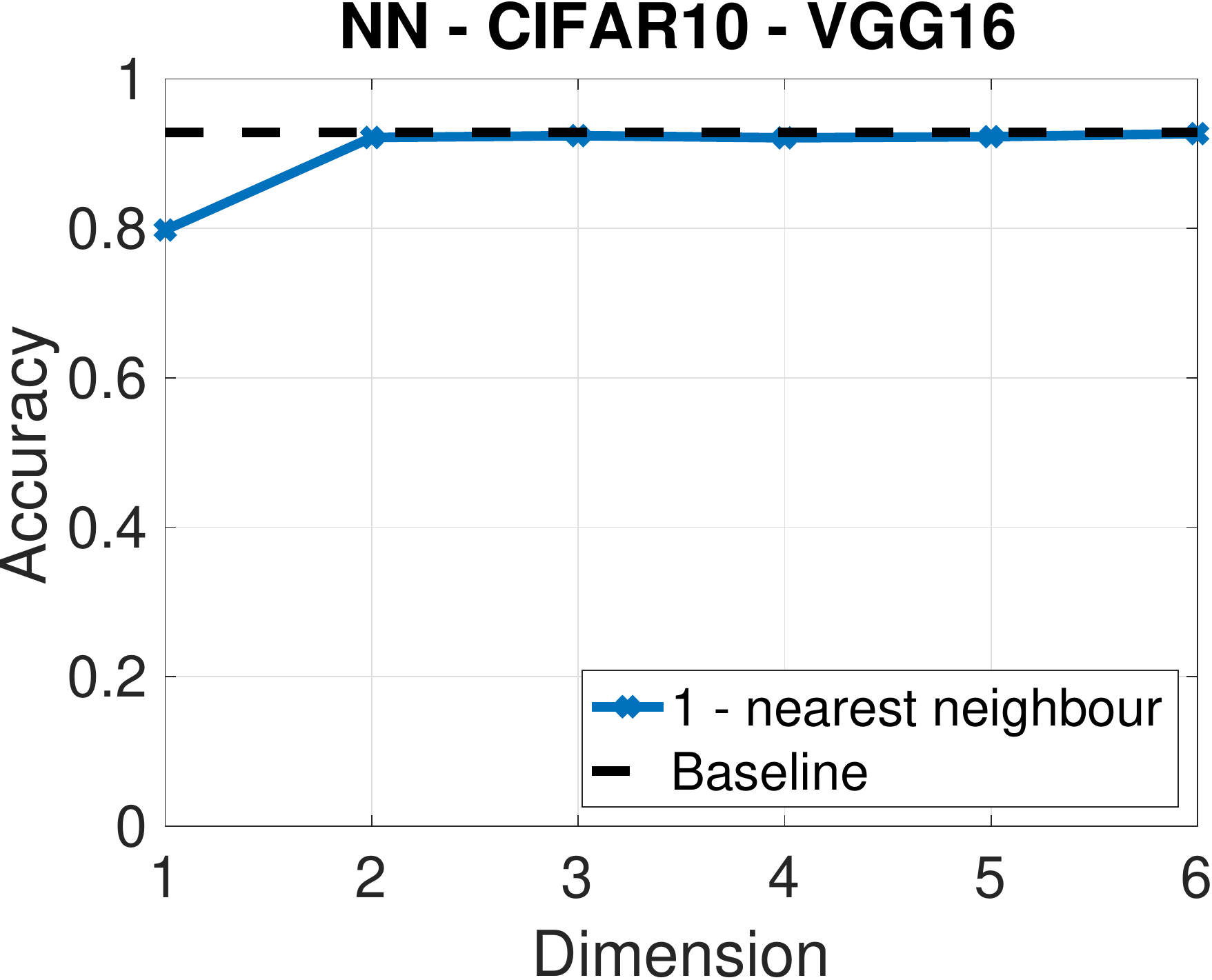}
    \includegraphics[width=0.32\textwidth]{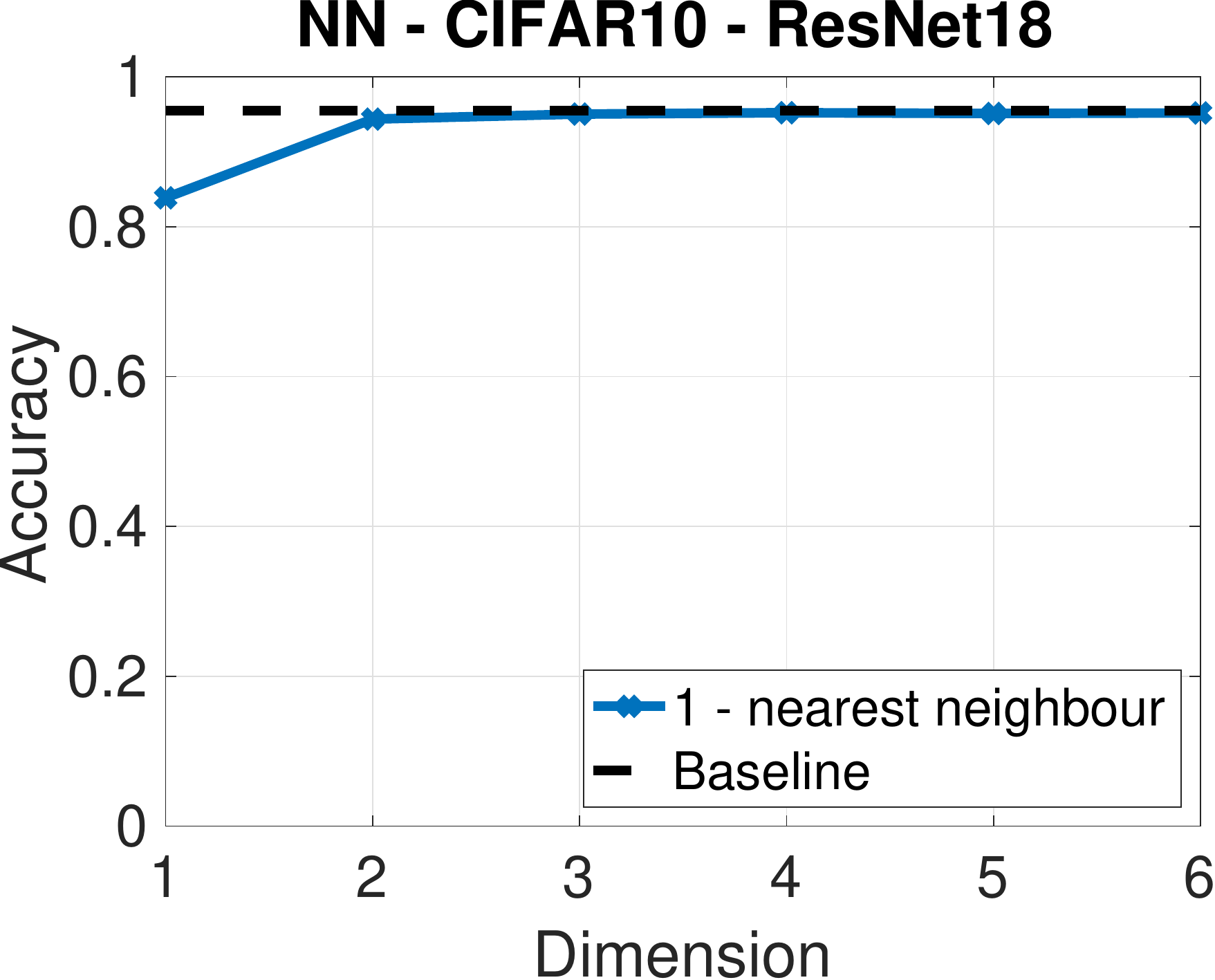}
    \includegraphics[width=0.32\textwidth]{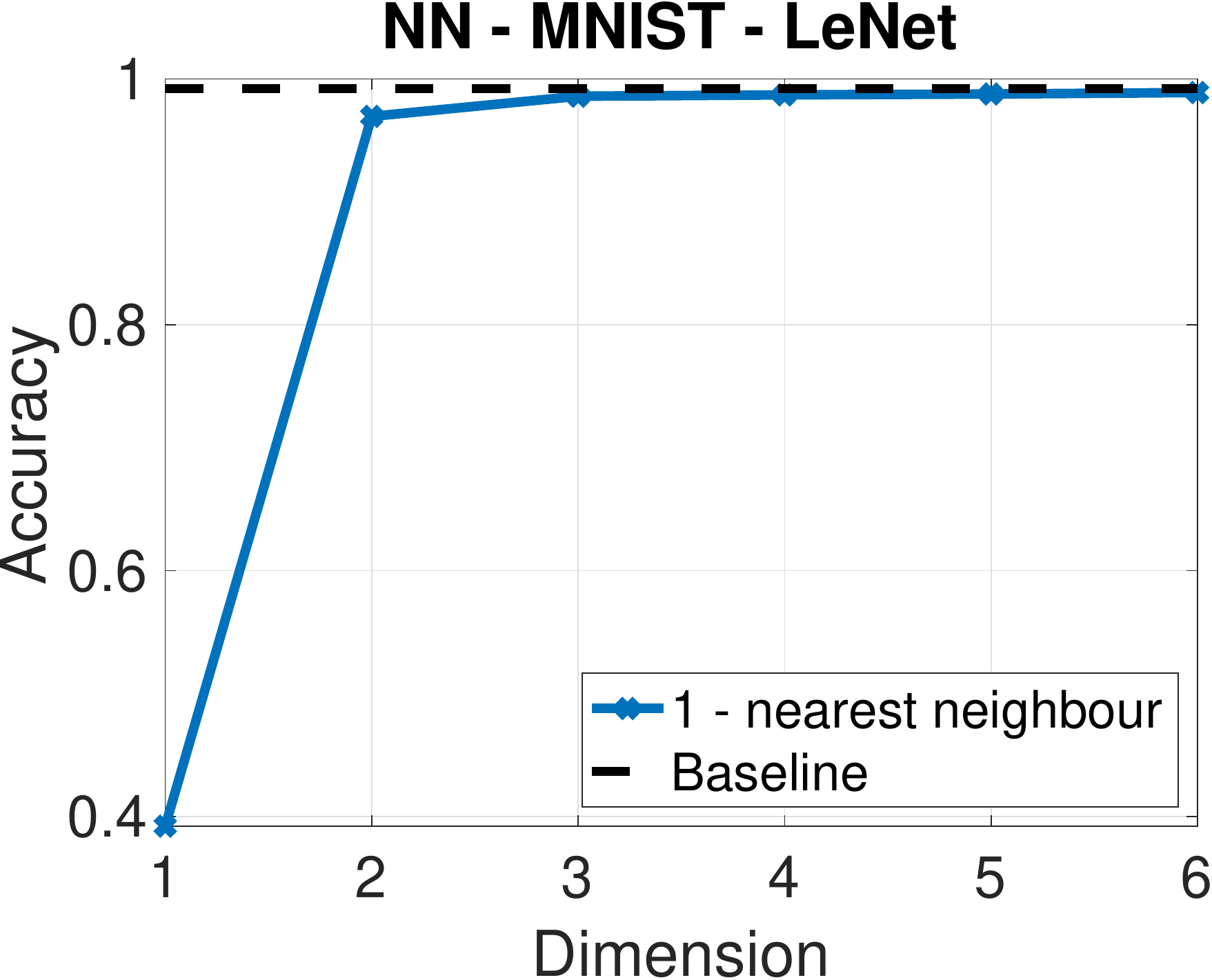} \\
    \includegraphics[width=0.32\textwidth]{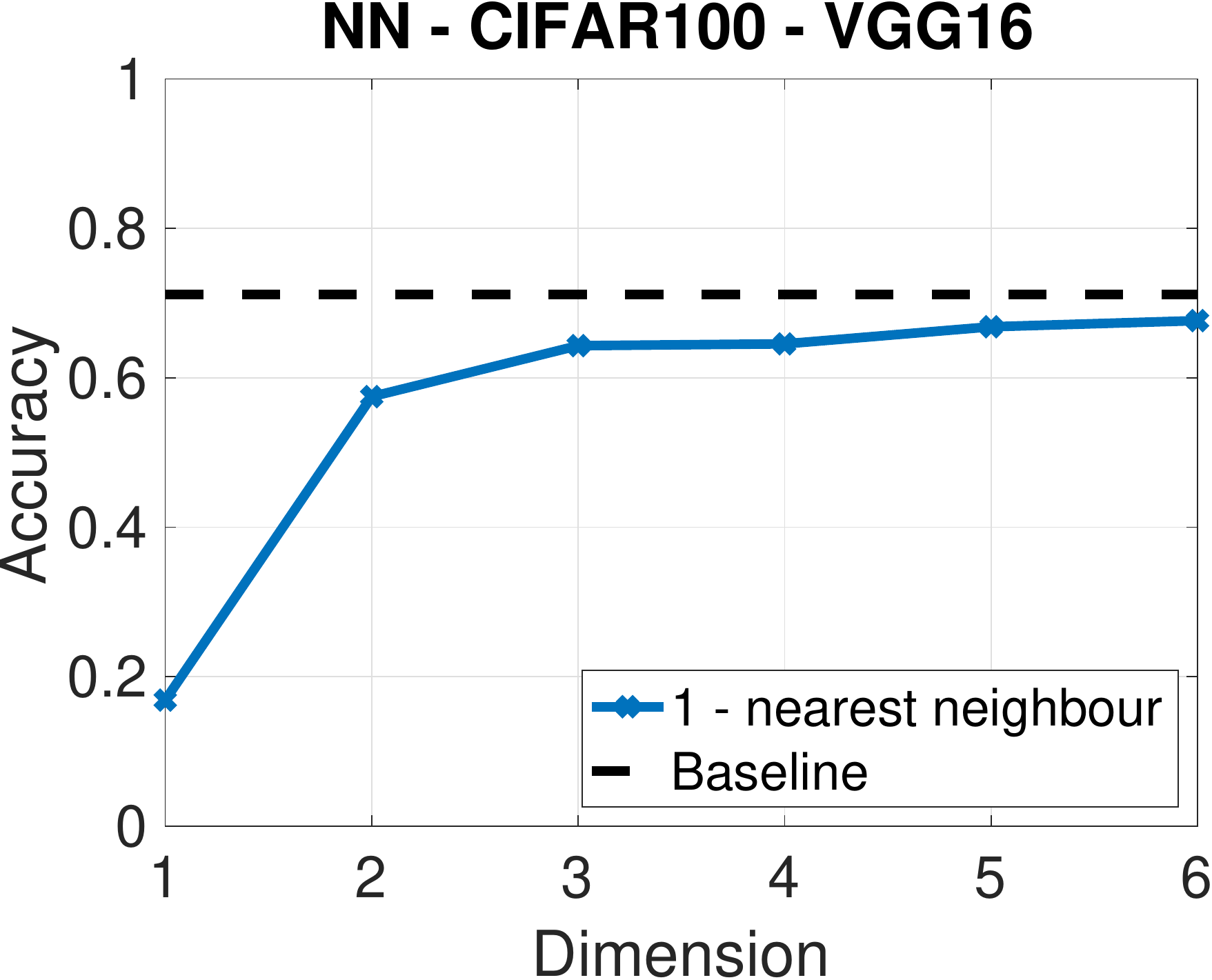}
    \includegraphics[width=0.32\textwidth]{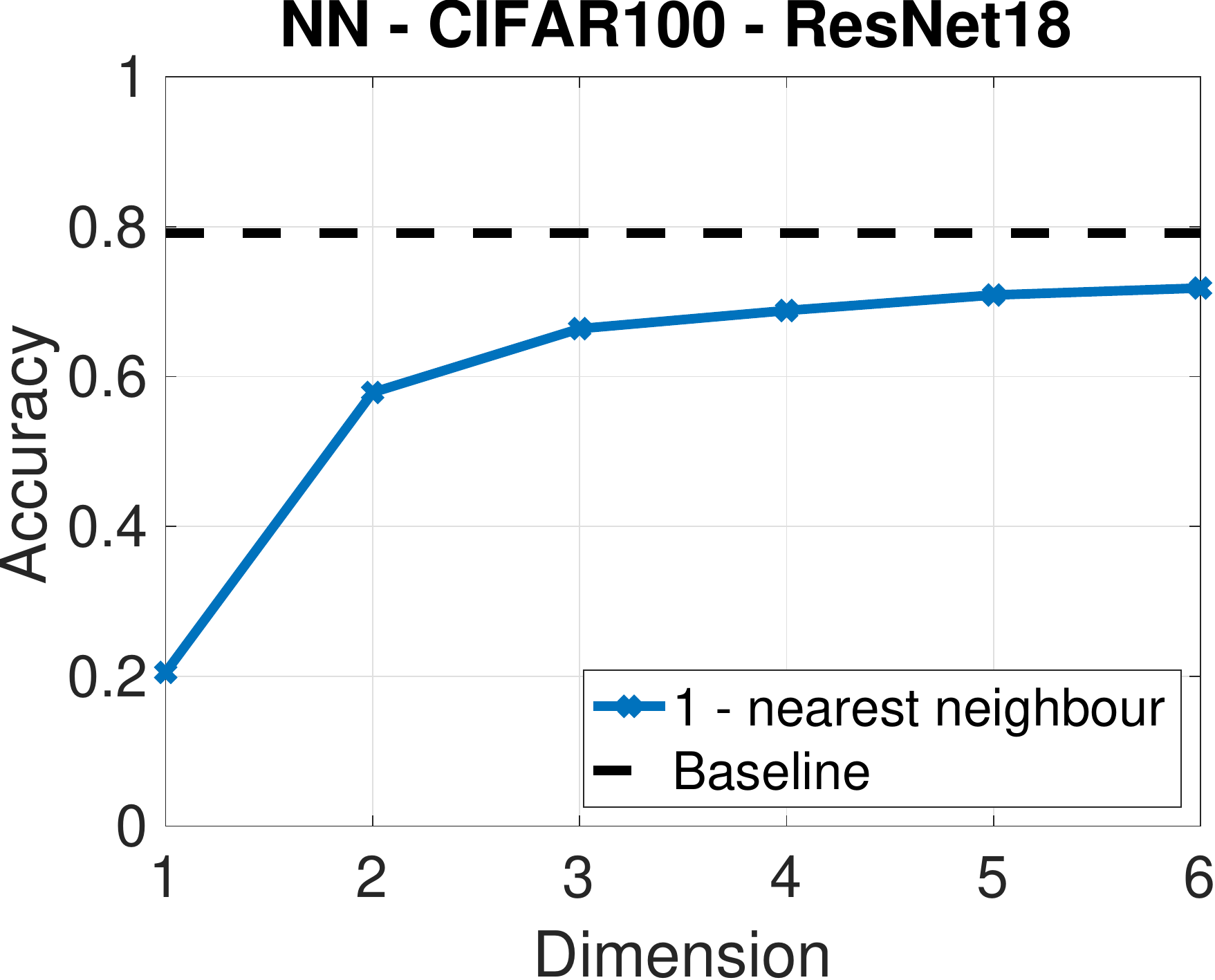}
    \includegraphics[width=0.32\textwidth]{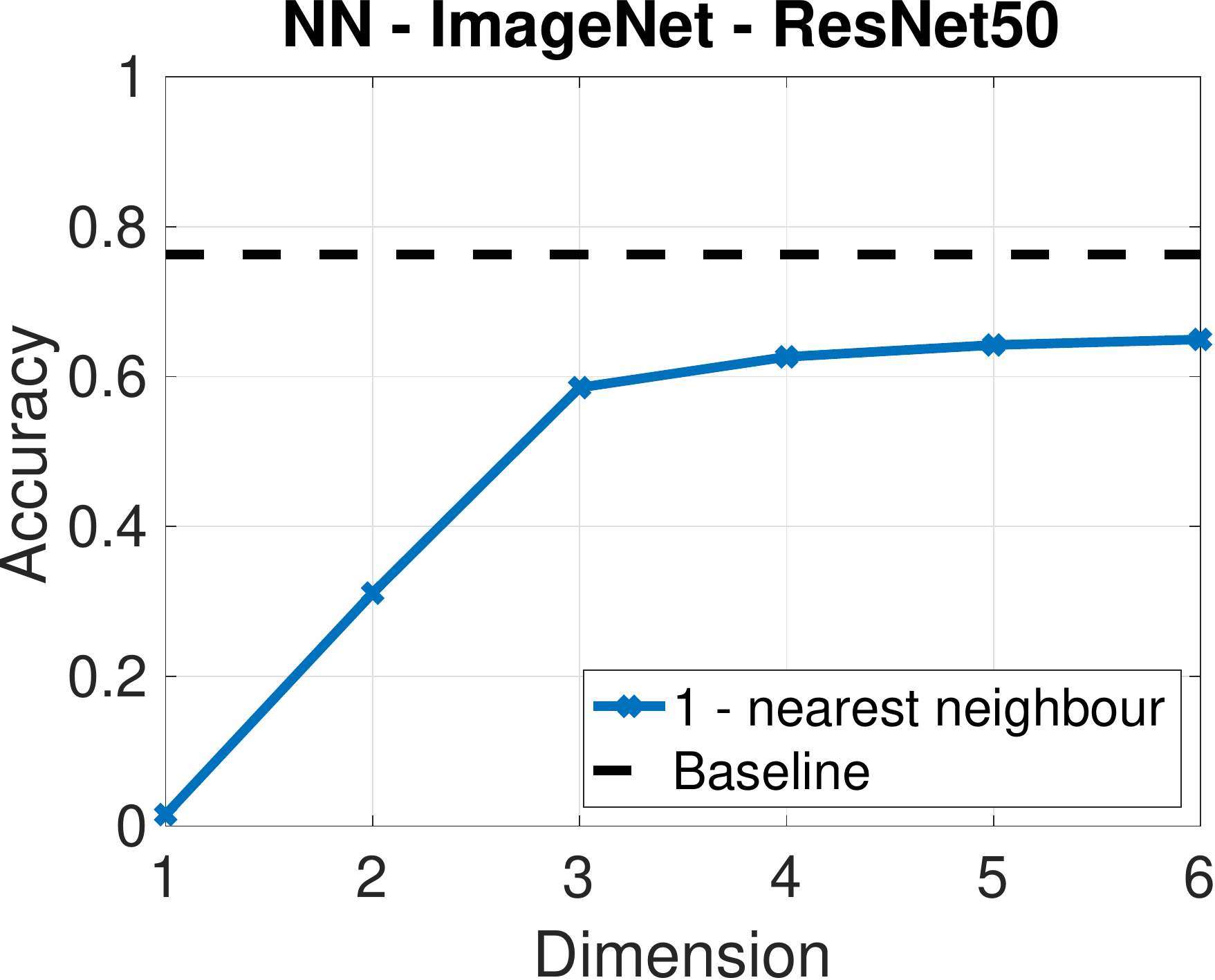}
    \caption{Nearest neighbor classification accuracy for different sizes of feature vectors ($d'=1,2,3,4,5,6$). Baseline model is the original DNN without bottleneck trained on full training data set and is marked with dashed line on the plots. \textbf{Top Row, from left to right:} VGG16 trained on CIFAR-10, ResNet-18 trained on CIFAR-10, and LeNet trained on MNIST, \textbf{Bottom Row, from left to right:} VGG16 trained on CIFAR-100, ResNet-18 trained on CIFAR-100, and ResNet-50 trained on ImageNet.}
    \label{fig:supp_knn}
\end{figure}

\subsection{Learning Curves}
\begin{table}[H]
    \centering
    \begin{tabular}{|l|l|l|c|c|c|p{1.6cm}|p{1cm}|}
        \hline
         Data & Model & Loss & BS & Opt           &  LR & LR decay ($\times 0.1$ at ep) & Weight decay \\[2ex] \hline  
         MNIST     & LeNet     & CE  & 128 & SGD  & 0.01   & -                        & 0.0  \\[0.4ex] \hline
         CIFAR-10  & ResNet18  & CE  & 128 & SGD  & 0.1    & [150, 250]  & $0.0005$  \\[0.4ex] \hline
         CIFAR-10  & VGG16     & CE  & 128 & SGD  & 0.01   & [150, 250]  & $0.0005$    \\[0.4ex] \hline
         CIFAR-100 & ResNet18  & CE  & 128 & SGD  & 0.1    & [150, 250]  & $0.0005$ \\[0.4ex] \hline
         CIFAR-100 & VGG16     & CE  & 128 & SGD  & 0.01   & [150, 250]  & $0.0005$   \\[0.4ex] \hline
         ImageNet  & ResNet-50 & CE  & 256 & SGD  & 0.1    & [30,60,90]  & $0.0001$ \\[0.4ex] \hline
         Udacity   & CovNet    & MSE & 64  & Adam & 0.0002 & [150]       & $0.0$ \\[0.4ex] \hline
    \end{tabular}
    \caption{Hyper-parameters of different experiments. CE - Cross Entropy, MSE - Mean Square Error, Opt - Optimizer and LR - Learning Rate.}
    \label{tab:my_label}
\end{table}

\clearpage 
\subsubsection{CIFAR-10}
\vspace{-0.3cm}

\subsubsection{CIFAR-100}

\subsubsection{ImageNet}

\subsubsection{MNIST}

\subsubsection{Udacity}
\begin{table}[H]
    \centering
    \begin{tabular}{|c|c|c|c|c|c|c|}
        \hline
        Layer & Output Size & Kernel & Stride  & Padding \\
        \hline
        Conv   & $32 \times 64 \times 64$  & $3\times3$ & 1 & 1  \\ \hline
        Conv   & $64 \times 32 \times 32$  & $3\times3$ & 1 & 1  \\ \hline
        Conv   & $128 \times 16 \times 16$ & $3\times3$ & 1 & 1  \\ \hline
        Conv   & $128 \times 8 \times 8$   & $3\times3$ & 1 & 1  \\ \hline
        Linear & $ 1024$ & - & - & - \\ \hline
        Linear & $ 1$ & - & - & - \\ \hline
    \end{tabular}
    \caption{CNN architecture used in Udacity experiments. Each Convolution layer is followed by ReLU, Maxpool, and Dropout layers and each Linear layer is followed by ReLU and Dropout except the last linear layer.}
    \label{tab:CovNet}
\end{table}

\begin{figure}[H]
    \centering
    \includegraphics[width=0.32\textwidth]{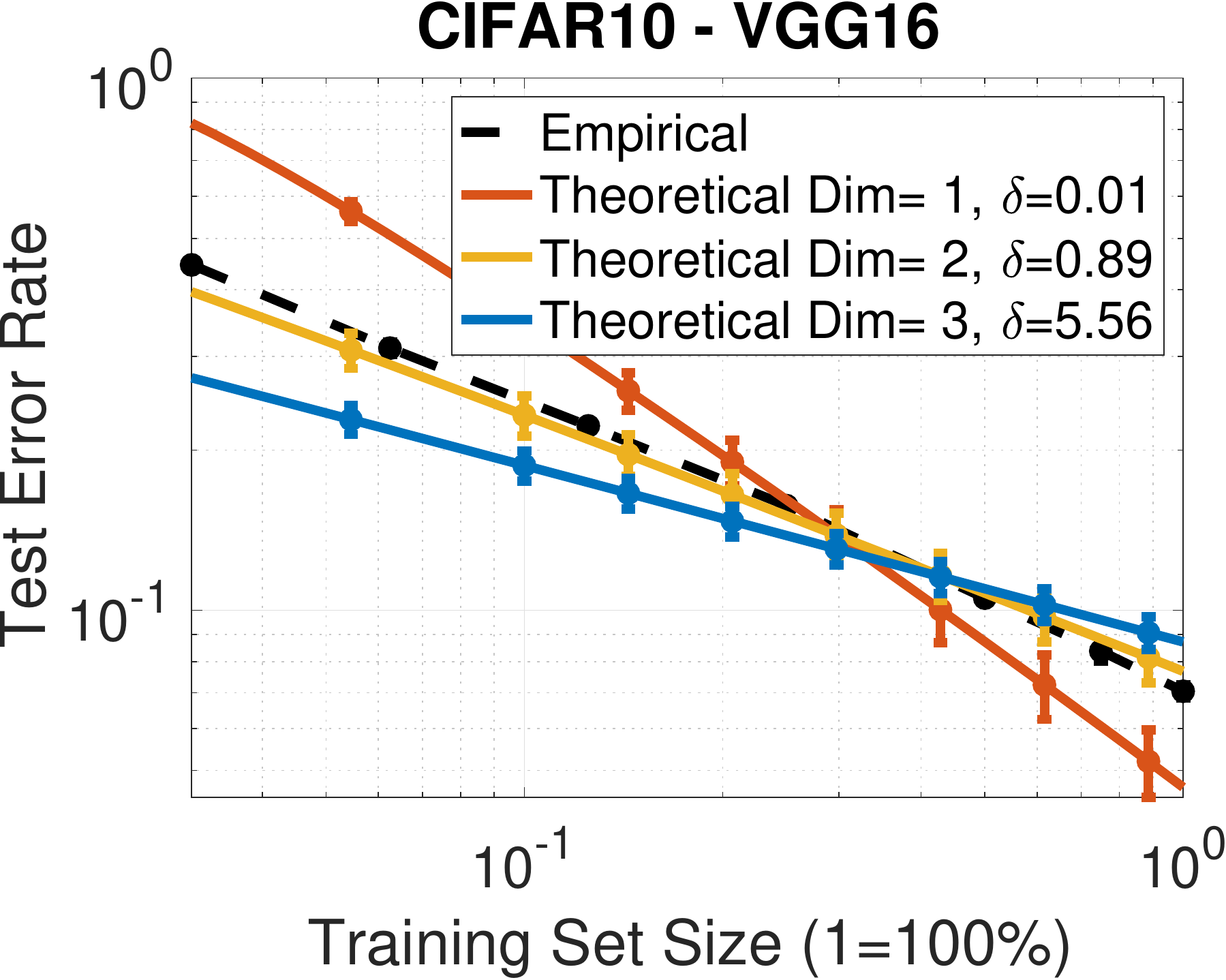}
    \includegraphics[width=0.32\textwidth]{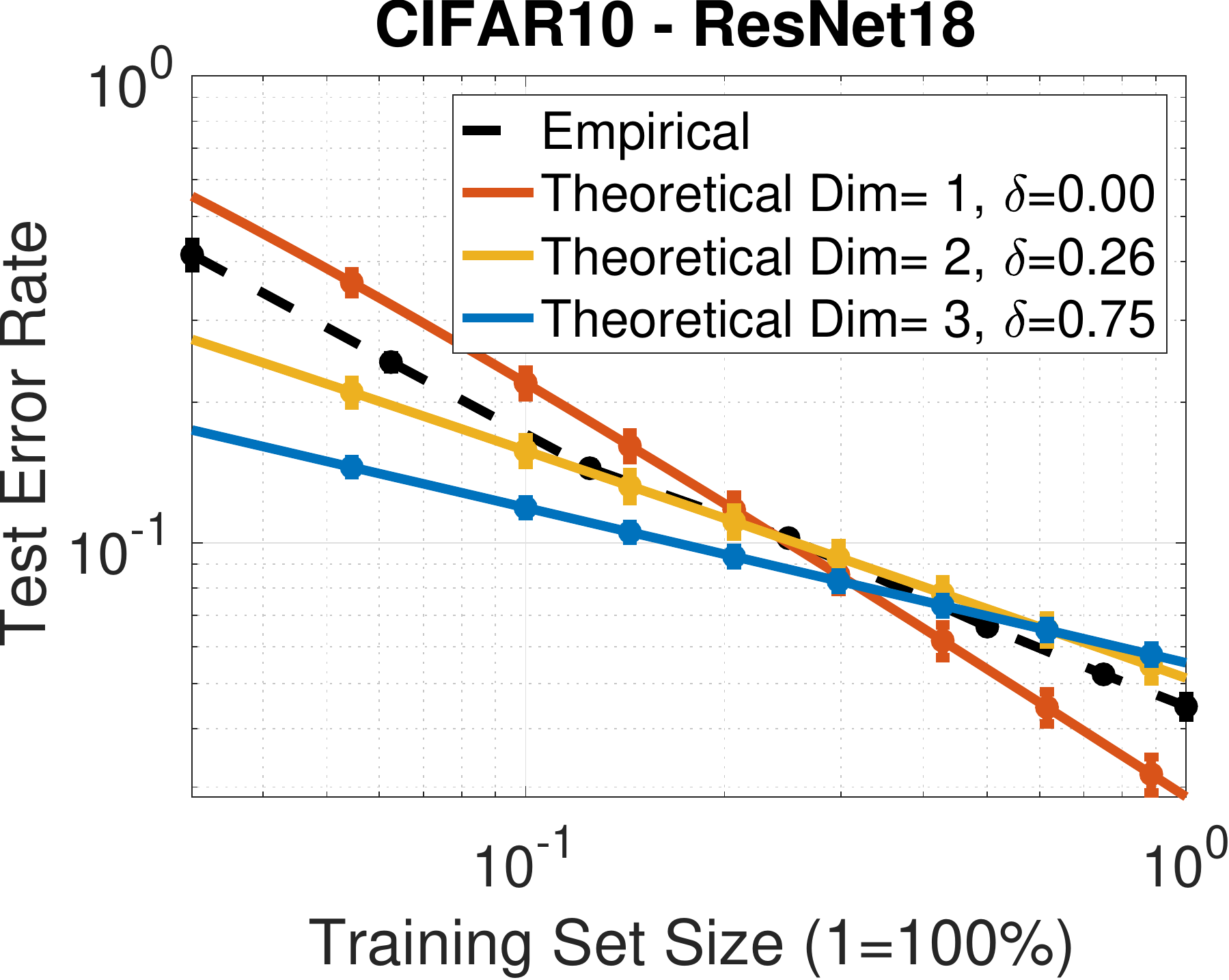}
    \includegraphics[width=0.32\textwidth]{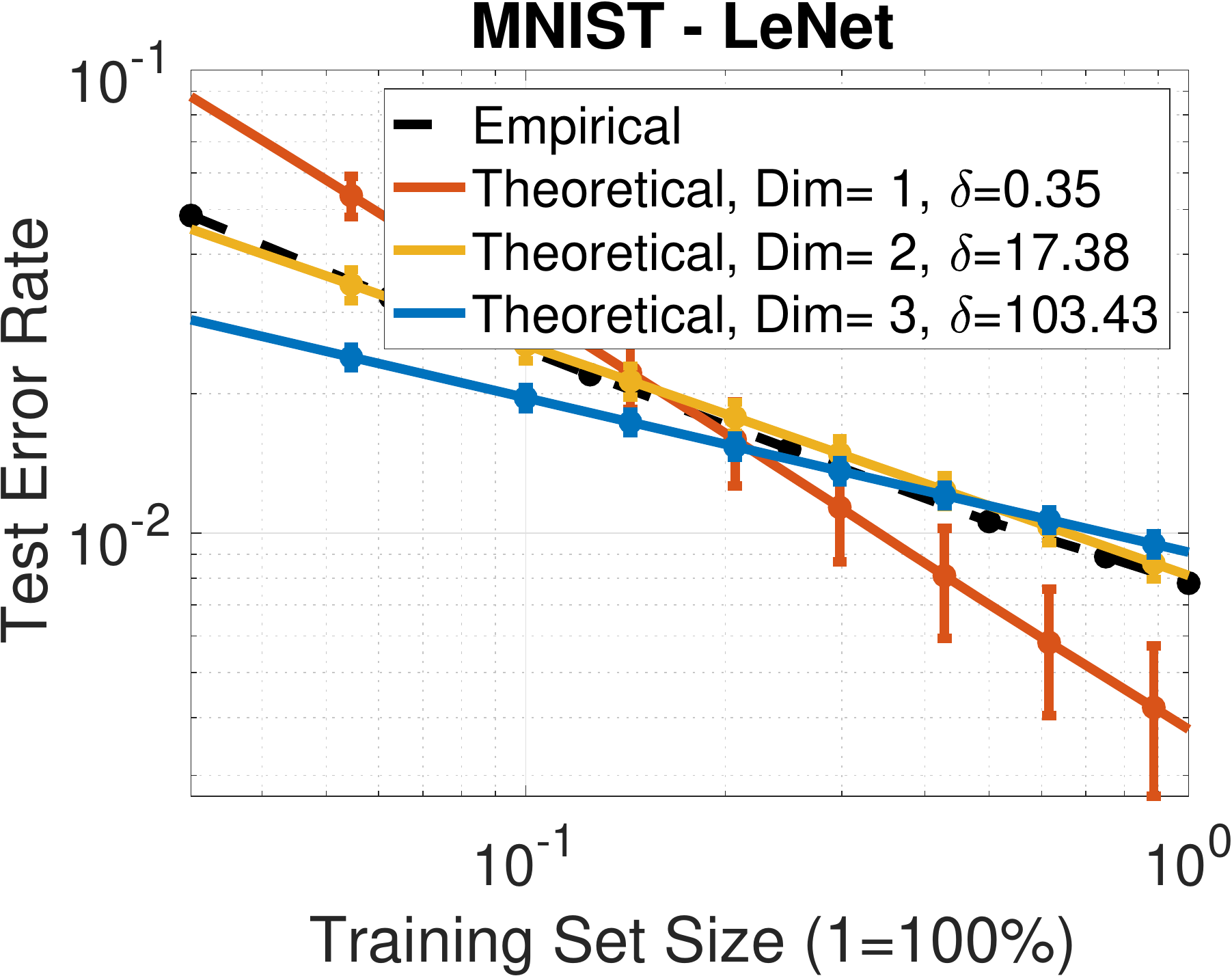}\\
    \includegraphics[width=0.32\textwidth]{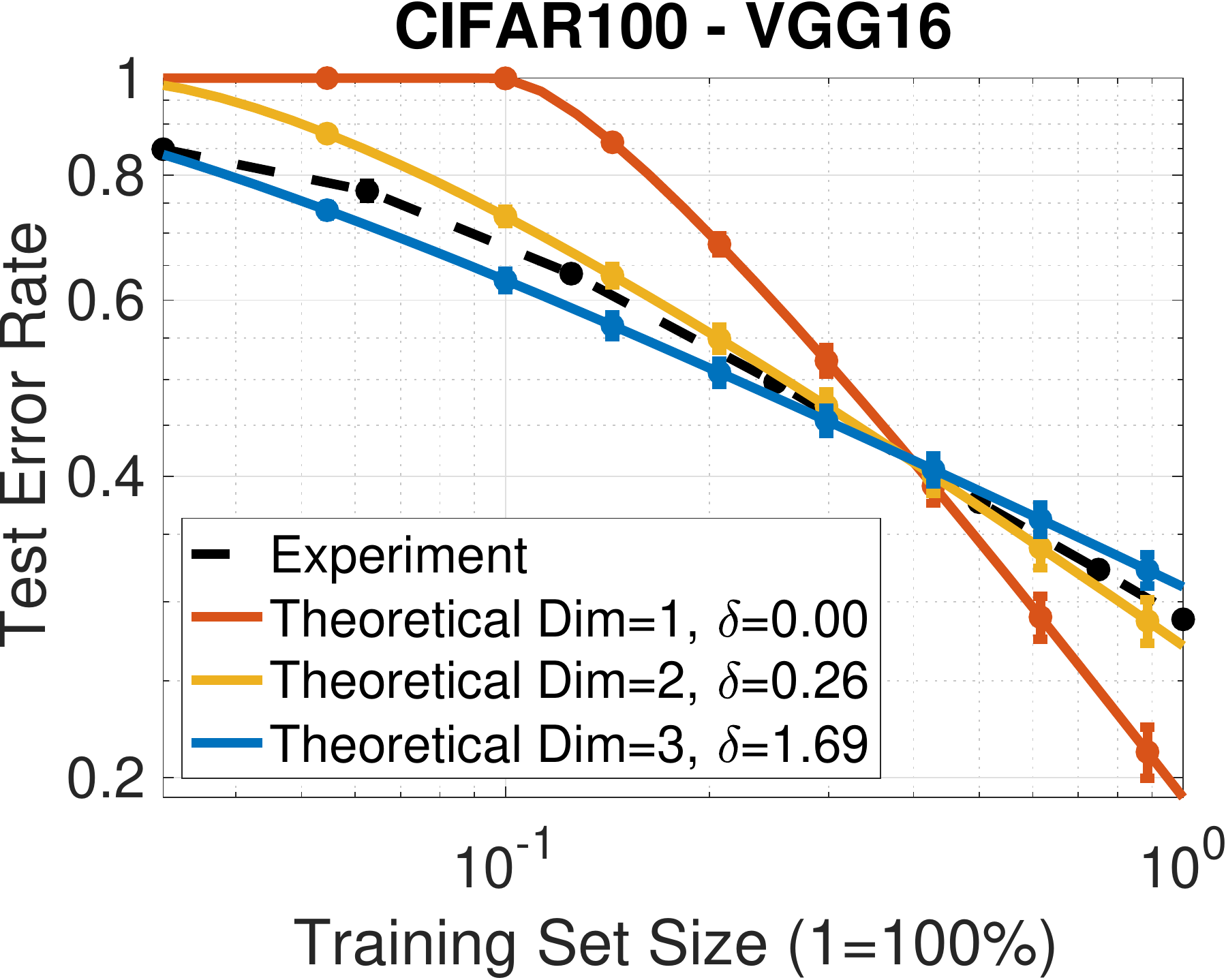}
    \includegraphics[width=0.32\textwidth]{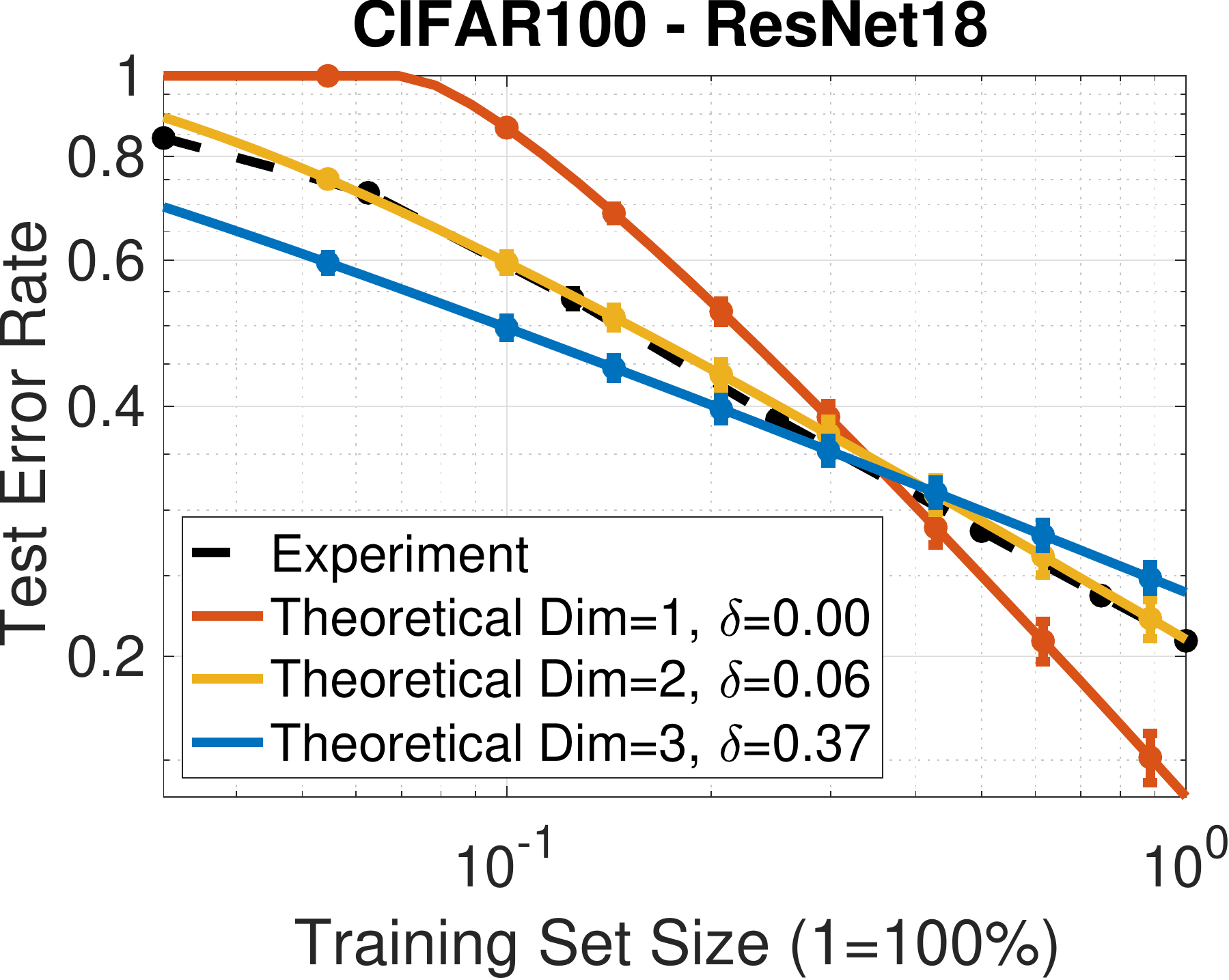}
    \includegraphics[width=0.32\textwidth]{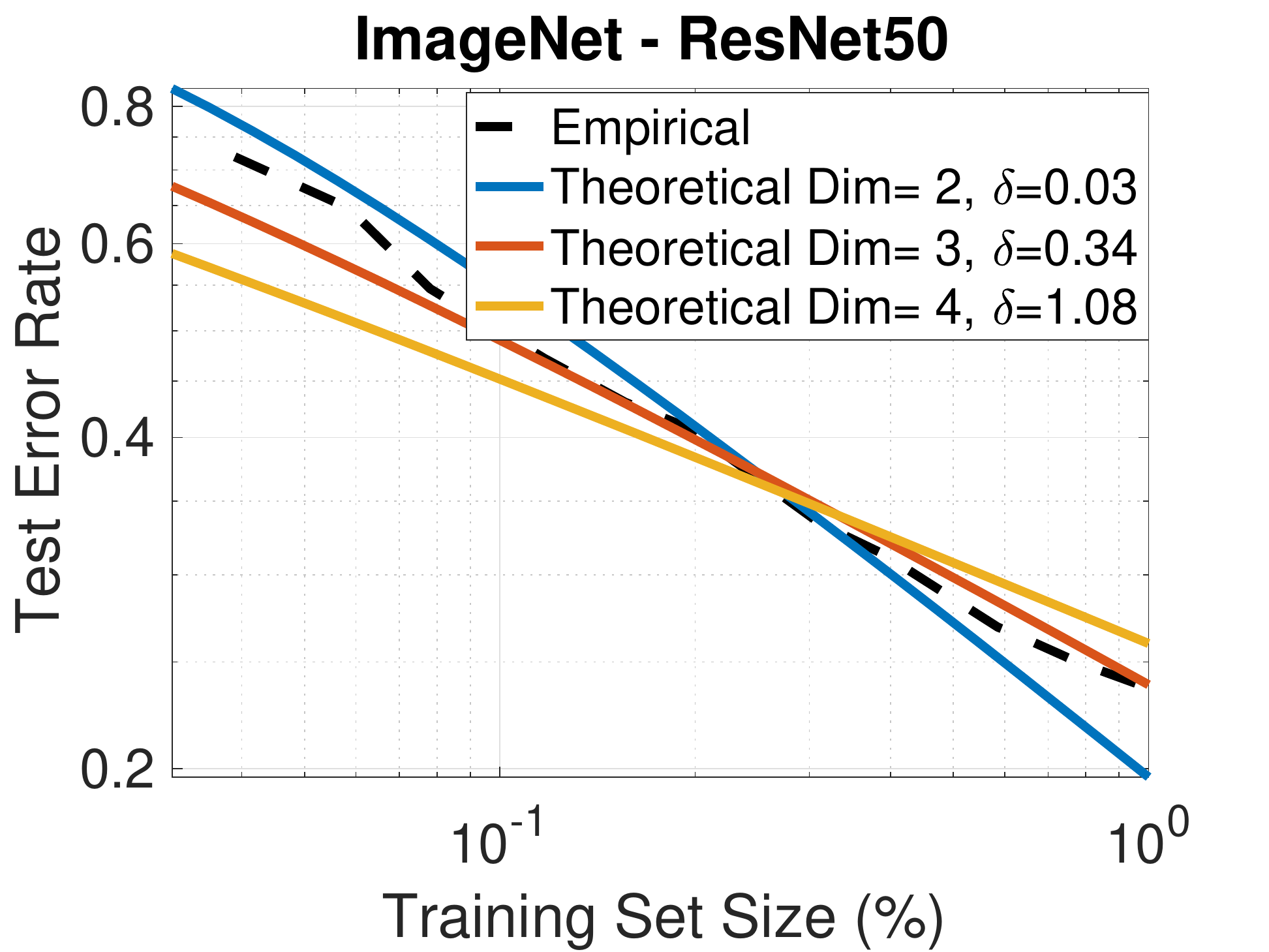} \\
    \includegraphics[width=0.32\textwidth]{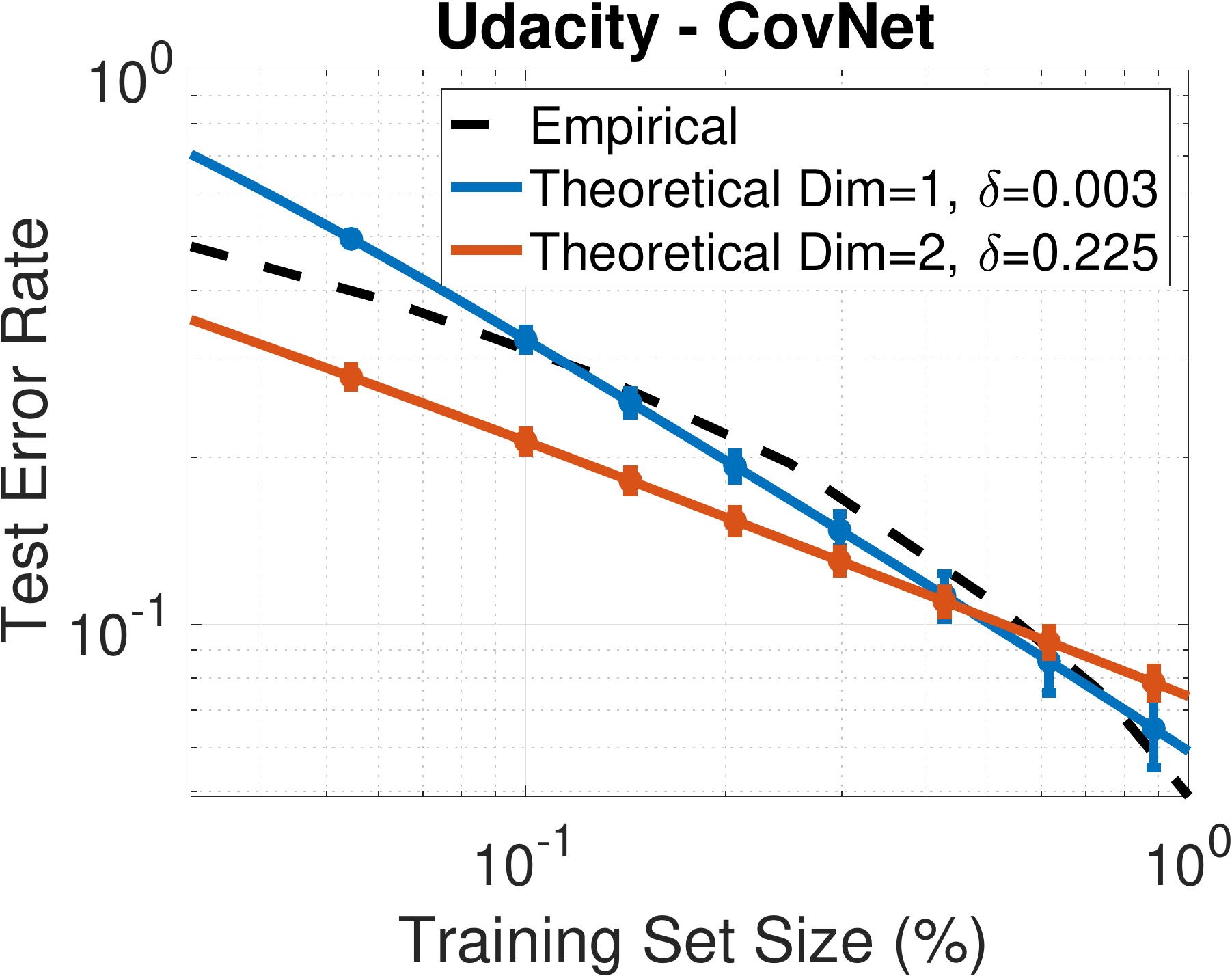}
    \caption{Empirical and theoretical learning curves (the latter obtained for different values of effective dimensionality).}
    \label{fig:supp_class_results}
\end{figure}

\end{document}